\documentclass[preprint,12pt,authoryear]{elsarticle}

\usepackage[left]{lineno}
\usepackage{amssymb}
\usepackage{amsmath}
\usepackage{graphicx}
\usepackage{enumitem}
\usepackage{booktabs}
\usepackage{multirow}
\usepackage{placeins} 
\usepackage{xcolor}
\usepackage{float}
\usepackage[authoryear]{natbib}
\usepackage{url}
\usepackage{hyperref}
\usepackage{threeparttable}
\usepackage{subcaption}
\usepackage{makecell}
\usepackage{tabularx}
\usepackage{adjustbox}
\usepackage[table]{xcolor}

\journal{Computer Environment and Urban Systems}

\begin{document}
% \linenumbers

\begin{frontmatter}

\title{UST-GNN: A Unified Spatial--Topological Graph Neural Network Framework for Urban Analytics \\ 
\large Demonstrated through a Case Study on Urban Health Prediction}
% \author{Anonymous authors}
\author[inst1,inst2]{Minwei Zhao}
\author[inst3,inst5]{Sanja \v{S}\'{c}epanovi\'{c}}
\author[inst1]{Stephen Law\corref{cor1}}
\ead{stephen.law@ucl.ac.uk}
\author[inst4]{Ivica Obadic}
\author[inst2]{Cai Wu}
\author[inst3,inst6]{Daniele Quercia}

\cortext[cor1]{Corresponding author.}

\affiliation[inst1]{organization={University College London},
            city={London},
            country={United Kingdom}}

\affiliation[inst2]{organization={The Hong Kong University of Science and Technology (Guangzhou)},
            city={Guangzhou},
            country={China}}

\affiliation[inst3]{organization={Nokia Bell Labs},
            city={Cambridge},
            country={United Kingdom}}

\affiliation[inst4]{organization={Technical University of Munich},
            city={Munich},
            country={Germany}}

\affiliation[inst5]{organization={University of Oxford},
            city={Oxford},
            country={United Kingdom}}

\affiliation[inst6]{organization={Politecnico di Torino},
            city={Turin},
            country={Italy}}

\begin{abstract}
Understanding how social, demographic, environmental, and spatial factors jointly shape urban outcomes is essential for sustainable urban development and evidence-based policy. Traditional statistical approaches often struggle to capture complex non-linear relationships, while many machine learning methods overlook the joint roles of spatial autocorrelation and network topology in urban systems. Recent advances in GeoAI have addressed these challenges only partially, often treating spatial effects, graph structure, evaluation, and interpretability separately. We present \textbf{UST-GNN}, a unified spatial--topological graph neural network framework that integrates neighbourhood connectivity, heterogeneous urban features, and positional/locational embeddings into a single representation. Using the MedSAT dataset, which contains over 150 environmental and socio-demographic variables and six prescription outcomes across 4,835 neighbourhoods in Greater London, UST-GNN outperforms strong statistical, geographically enhanced, and graph Machine Learning baselines, improving out-of-sample $R^2$ by 8.4--13.2\% under strict spatial cross-validation. We further introduce a lightweight principal-component module to interpret learned node embeddings geographically and relate them to policy-relevant covariates. The resulting analyses recover established patterns, offer new perspectives on debated associations, and reveal novel predictors warranting further causal investigation. Together, these findings demonstrate the value of graph-based spatial machine learning for urban health analytics, environmental inequality assessment, and evidence-based urban policy. Beyond predictive gains, UST-GNN provides a unified GeoAI analytical pipeline that can be embedded into urban digital twin workflows for scenario testing, monitoring, and data-informed decision-making for healthier, more sustainable cities.
\end{abstract}


\end{frontmatter}

\section{Introduction}
Urban outcomes, including health, are shaped by the complex interplay of socio-demographic, environmental, and infrastructural factors, making their systematic study essential for sustainable urban development and informed public policy. Recent global trends---such as rapid urbanisation \citep{vlahov2002urbanization}, the increasing frequency and severity of extreme weather events \citep{bell2018changes}, and pandemics such as COVID-19 \citep{capolongo2020covid}\allowbreak{} have heightened the urgency of understanding how urban characteristics influenced population health and contributed to existing inequalities \citep{fiscella2004health, cole2021breaking}. As cities worldwide strove to meet the United Nations Sustainable Development Goals (SDGs), particularly those related to good health and wellbeing (SDG~3) and sustainable cities and communities (SDG~11), policymakers and planners increasingly required systematic, spatially explicit, and evidence-based insights to guide equitable urban planning and targeted interventions \citep{obradovich2018empirical, corburn2004confronting, batty2018artificial}.

Urban and public health studies have traditionally employed statistical methodologies \citep{alegria2018social, hayat2017statistical, cousens2011alternatives} and, more recently, machine learning (ML) approaches \citep{dos2019data, scepanovic2024medsat, maitra2025location} to explore relationships between the urban environment and health outcomes. Recent advances in Artificial Intelligence (AI) and Digital Twin technologies are increasingly transforming urban analytics by coupling data-driven models with dynamic, real-time city representations. This convergence enables continuous simulation and feedback between physical and virtual urban systems, supporting proactive responses to health, mobility, and environmental challenges~\citep{birks2020towards,malleson2024digital}. Integrating GeoAI models into digital-twin frameworks offers new opportunities for evidence-based planning and scenario evaluation toward healthy and sustainable cities.

Despite these advances, important methodological limitations remain. First, conventional statistical models often assumed linear relationships for interpretability, limiting their ability to capture complex non-linear interactions among environmental exposures, socio-economic conditions, and other urban factors \citep{song2004comparison, pineo2018urban, lalloue2013statistical, kihal2013green}. Second, many machine learning approaches overlooked the simultaneous spatial dependencies and network topology inherent in urban systems, failing to account for neighbourhood severance and structural linkages such as transport or social networks that may propagate risks \citep{arcaya2016research, sarkar2017urban}. These omissions can reduce predictive accuracy and hinder understanding of spatial disparities \citep{ijeh2024predictive}. Third, although location encoder has improved the representation of spatial context in machine learning by embedding geographic information directly into feature space \citep{klemmer2023satclip, mai2020multi}, such methods remain rarely integrated with architectures that jointly model spatial proximity and topological connectivity in urban health analytics. Graph Neural Networks (GNNs) offer a promising avenue for this challenge because they are well suited to graph-structured data and can capture complex spatial-topological relationships, with demonstrated success in domains including social networks, molecular chemistry, recommender systems, and, more recently, urban informatics tasks such as traffic prediction and urban function classification \citep{li2023survey, sun2022gnn, wu2022graph, liu2024association, zhang2023knowledge, gao2024uncertainty}. However, their application to urban health remains limited, partly because their black-box nature constrains uptake among public health researchers and policymakers \citep{liu2024explainable, kjellstrom2016impact}. This is especially important as urban and public health research increasingly calls for machine learning and AI methods with strong predictive performance \citep{mhasawade2021machine, kong2023leveraging, gilardi2021social}. Existing post hoc explainability methods, such as GNNExplainer and gradient-based attribution, can identify influential nodes, edges, and features, but they typically provide instance-specific explanations rather than global, policy-relevant concepts for urban analytics.

In summary, recent GeoAI advances have largely addressed these geospatial challenges in isolation, modelling spatial effects through location encoders, topological dependencies through graph-based learning, spatial leakage through spatial cross-validation, and interpretability through post hoc explanation frameworks such as GNNExplainer. Few studies have unified or systematically ablated these components within a single spatially explicit GeoAI pipeline designed to generate health-policy-relevant insights. To address this gap, we propose \textbf{UST-GNN}, a unified spatially and topologically explicit graph neural network framework for urban analytics, demonstrated here through a case study on health outcomes. UST-GNN combines position/location encoders, neighbourhood connectivity, and diverse urban features within a unified graph representation to model interactions among population, environmental, and spatial variables. Using a spatial graph constructed from neighbourhood-level data, it integrates heterogeneous features, including air quality, greenness, climate, and population measures, together with positional and locational embeddings, to predict neighbourhood-level prescription outcomes within an ecological study design. UST-GNN also introduces a lightweight interpretability module that summarizes the principal components of the learned representation, identifies spatial patterns, and links these components back to policy-relevant input features. We show that UST-GNN outperforms state-of-the-art baselines under strict spatial cross-validation while generating interpretable insights that align with, challenge, and extend current urban health knowledge. Our contributions are threefold:

\begin{enumerate}
    \item We developed \textbf{UST-GNN}, a unified spatial-topological graph neural network pipeline that explicitly captured geographic dependencies and neighbourhood topology for urban analytics. Through systematic spatial ablation analyses, we rigorously evaluated alternative graph construction strategies (e.g., $K$-hop adjacency, $K$-nearest neighbours), diverse positional and locational encoding techniques (e.g., Laplacian, random-walk positional encodings, and SatCLIP locational encodings~\citep{klemmer2023satclip}), multiple GNN architectures (GCN, GIN, GraphSAGE, GATv2) and multiple spatial cross validation strategies. This established a spatially-explicit  methodological pipeline for integrating graph machine learning into urban analytics research, including but not limited to health applications (Section~\ref{sec:res_ablation}).
    \item We applied UST-GNN to the MedSAT dataset~\citep{scepanovic2024medsat}, a recently released urban health resource compiling over 150 high-resolution environmental and socio-demographic indicators, together with six health outcome measures (per-capita prescription rates for depression, anxiety, diabetes, hypertension, asthma, and opioid use) across 4,835 neighbourhoods in Greater London. UST-GNN achieved an average error reduction exceeding 10\% compared to competitive baselines, outperforming gradient-boosted trees, Spatial Lag Models, Geographically Weighted Regression, and geo-informed machine learning methods~\citep{scepanovic2024medsat, maitra2025location,grekousis2025geographical}. These results confirmed UST-GNN's ability to capture complex non-linear effects and spatial topological effects that were often overlooked by conventional spatial statistical models and standard ML approaches (Section~\ref{sec:res_performance}).
    \item Using antidepressant prescription rates as a representative case study, we developed a novel lightweight module that integrates principal component analysis (PCA) with spatial and socio-economic correlation analysis. The module compresses the learned graph representations into two intrinsic components, maps them geographically to reveal spatial structure, and relates each component back to health-policy-relevant input features. Antidepressants were selected because they have been extensively studied in prior work, enabling comparison with the existing literature. Unlike commonly used interpretability techniques, such as feature permutation, gradient based attribution, or GNNExplainer (see~\ref{app:explainability}), which primarily operate at the input or node level, our approach interprets the dominant component and its global associations with policy-relevant input features, which are often overlooked by local perturbation or feature-wise methods. By relating the intrinsic components back to the original input features, our framework recovered socio-economic and demographic relationships consistent with findings from more interpretable models in health research (e.g., lower prescriptions among individuals aged 20--39, commuters, and those in professional employment), provided new perspectives on debated associations (e.g., the paradoxical relationship between increased urban greenery and higher antidepressant use), and identified novel spatial predictors (e.g., a negative correlation between canopy evaporation and depression prescriptions) that may warrant further causal investigation (Section~\ref{sec:res_XAI}). Discovering such health-policy-relevant insights from a complex GeoAI model represents a key contribution of this study, demonstrating how advanced spatial deep learning architectures can be linked to actionable evidence for urban health decision-making.
\end{enumerate}

\section{Literature Review}
\subsection{Urban socio-demographic and environmental correlates of outcomes}

A substantial body of research has examined how urban socio-demographic and environmental characteristics are linked to a range of outcomes, particularly public health and related inequalities. In the health domain, \citet{kihal2013green} demonstrated that low-income populations residing in environmentally disadvantaged neighbourhoods faced compounded risks, especially in contexts of pronounced social inequities. Similarly, access to urban green and blue spaces has been consistently associated with improved health outcomes; for example, \citet{zhou2023combined} found that increased availability of green spaces and waterways significantly mitigated hypertension risks. Other studies identified strong associations between urban environmental conditions and chronic health outcomes \citep{hunter2023advancing}, including depression and diabetes \citep{white2021associations, mazumdar2021green, van2020socioeconomic}. Beyond health, socio-demographic and environmental factors have also been linked to variations in mobility, environmental quality, and social wellbeing \citep{gao2022regional,yue2024substantially}.
Collectively, these findings underscored the critical role that environmental exposures and socio-economic contexts played in shaping urban disparities. However, many existing studies tended to adopt a relatively narrow analytical scope, often investigating single factors in isolation rather than addressing the broader, multifaceted urban context. As a result, the complex interactions and potential non-linear effects among multiple urban characteristics remained insufficiently explored, limiting comprehensive understanding of both urban health and other spatially distributed outcomes.

\subsection{Machine learning applications in urban analytics and health research}

The advent of extensive, high-dimensional urban datasets, including those derived from remote sensing, wearable technology, crowdsourced platforms, and administrative records, has accelerated the adoption of machine learning (ML) techniques in urban analytics, with public health emerging as a key application domain \citep{ohanyan2022associations}. Traditional statistical models, while valuable, are often constrained by assumptions of linearity and independence among variables, which limits their ability to capture the complex, non-linear interactions common in urban health contexts \citep{rothenberg2014flexible, shane2000urban, song2004comparison, pineo2018urban}. For instance, \citet{lalloue2013statistical} illustrated that linear models struggled to account for intricate urban health patterns arising from interdependent variables. In contrast, contemporary ML methods including random forests, gradient boosting, and neural networks offered enhanced flexibility, automatically capturing complex interactions and non-linearities inherent to diverse urban datasets \citep{jordan2015machine}. Recent studies have successfully employed such algorithms to predict diverse health outcomes, including depression incidence, diabetes prevalence, and hypertension risks, based on multifaceted urban data sources \citep{bhakta2016prediction, su2021use, thotad2023diabetes}.

Nevertheless, conventional ML approaches typically neglected the geographic context essential for accurately modelling community-level phenomena \citep{arcaya2016research, sarkar2017urban}. Standard ML treated spatial units, such as neighbourhoods, as independent entities, overlooking spatial heterogeneity and dependencies. Factors such as the influence of adjacent areas, the presence of spatial barriers (e.g., rivers or highways), and spatially structured socio-economic gradients could significantly impact health and other urban outcomes, yet were frequently ignored \citep{comber2011spatial,shrestha2016environmental}. While spatial statistical methods, including spatial lag models and geographically weighted regression (GWR), explicitly accounted for spatial interactions, they often exhibited limitations in scalability and in capturing non-linear spatial relationships in high-dimensional contexts \citep{cheng2014spatiotemporal,zhao2025characterization}. Moreover, the limited interpretability of many ML models remained an ongoing challenge. In urban health applications, understanding why particular regions exhibited elevated risks was as critical as achieving accurate predictions, in order to inform targeted and effective interventions \citep{rudin2019stop,caruana2015intelligible}. The lack of transparency inherent in many ``black box'' models thus constrained their acceptance and broader use in both public health and other urban policy domains \citep{wirtz2021artificial}.

\subsection{GeoAI and graph neural networks in urban analytics and health research}
Recognizing the significance of spatial context in urban studies, the emerging field of Geographic Artificial Intelligence (GeoAI) has integrated geographical information science (GIScience) with advanced machine learning and AI methodologies \citep{mai2020multi, vopham2018emerging, scheider2023pragmatic,klemmer2023satclip,wang2025multi}. GeoAI approaches have shown considerable potential in modelling location-based factors and spatial relationships, with applications spanning public health, environmental monitoring, and mobility analysis. In the health domain, research has underscored how urban spatial heterogeneity—including variations in built and social environments—contributed to disparities in health outcomes \citep{tung2017spatial}. The COVID-19 pandemic further highlighted the critical role of spatial context, prompting greater attention to neighbourhood characteristics and mobility patterns in analysing disease transmission and health impacts \citep{krenz2023linking, li2023survey}.

Within the GeoAI landscape, Graph Neural Networks (GNNs) have emerged as a particularly promising approach, capable of naturally capturing network topology and spatial relationships. In such frameworks, urban areas can be explicitly represented as interconnected nodes, with edges encoding spatial proximity, infrastructure connectivity, or socio-economic interactions \citep{klemmer2023positional, li2021prediction, wu2021inductive}. GNN-based methods have been successfully applied to complex phenomena such as human mobility modelling and infectious disease dynamics, both of which involve pronounced network effects \citep{choi2020learning}. \citet{helbich2018toward} highlighted the value of integrating diverse spatial datasets (e.g., environmental, infrastructural, socio-economic) to explain disparities in urban outcomes, a task well suited to graph-based approaches given their ability to model heterogeneous, relational data structures \citep{desabbata2023graph, mai2020multi}.

Despite this promise, the use of GNNs in urban health research remained limited, in large part due to challenges of interpretability. The complexity of GNN architectures often obscured intuitive understanding, posing a barrier for domain experts who require model reasoning to align with established theoretical and empirical knowledge. Existing applications of GNNs to health contexts, such as those by \citet{lu2021weighted} and \citet{fritz2022combining}, have largely emphasised predictive accuracy or epidemiological modelling, rather than describing the underlying socio-environmental drivers of health disparities. As a result, the potential of GNNs to combine predictive performance with domain-relevant interpretability in urban health analytics has remained underexplored. Recent post hoc explainability techniques, including GNNExplainer and gradient-based attribution, seek to isolate the minimal set of nodes, edges, and features that most influence a model’s prediction. However, these approaches largely explain the black box after training via saliency masks that are instance-specific rather than interpreting global policy-relevant concepts for health urban analytics. 

To address this gap, we proposed \textbf{UST-GNN}, a unified spatial and topologically explicit GNN framework designed for urban analytics, demonstrated here through a case study on health outcomes (per capita medical prescription distribution). UST-GNN combines robust predictive capabilities with interpretable spatial insights, enabling the identification of neighbourhood-level patterns and feature associations relevant in health policies. This approach aligns with core GeoAI objectives, providing a pathway to jointly address the intertwined challenges of modelling non-linear relationships, capturing spatial dependencies, and delivering interpretable outputs to support sustainable urban research and policy holistically. Moreover, such graph-based GeoAI approaches can form the analytical backbone of emerging urban digital twins, integrated virtual representations of cities, that enable continuous monitoring, simulation, and policy testing~\citep{malleson2024digital}. Embedding models like UST-GNN could facilitate dynamic assessment of urban health and sustainability scenarios, supporting real-time, data-informed decision-making for healthier and more sustainable cities.

\section{Methodology}
\begin{figure}[!htbp]
    \centering
    \includegraphics[width=\textwidth]{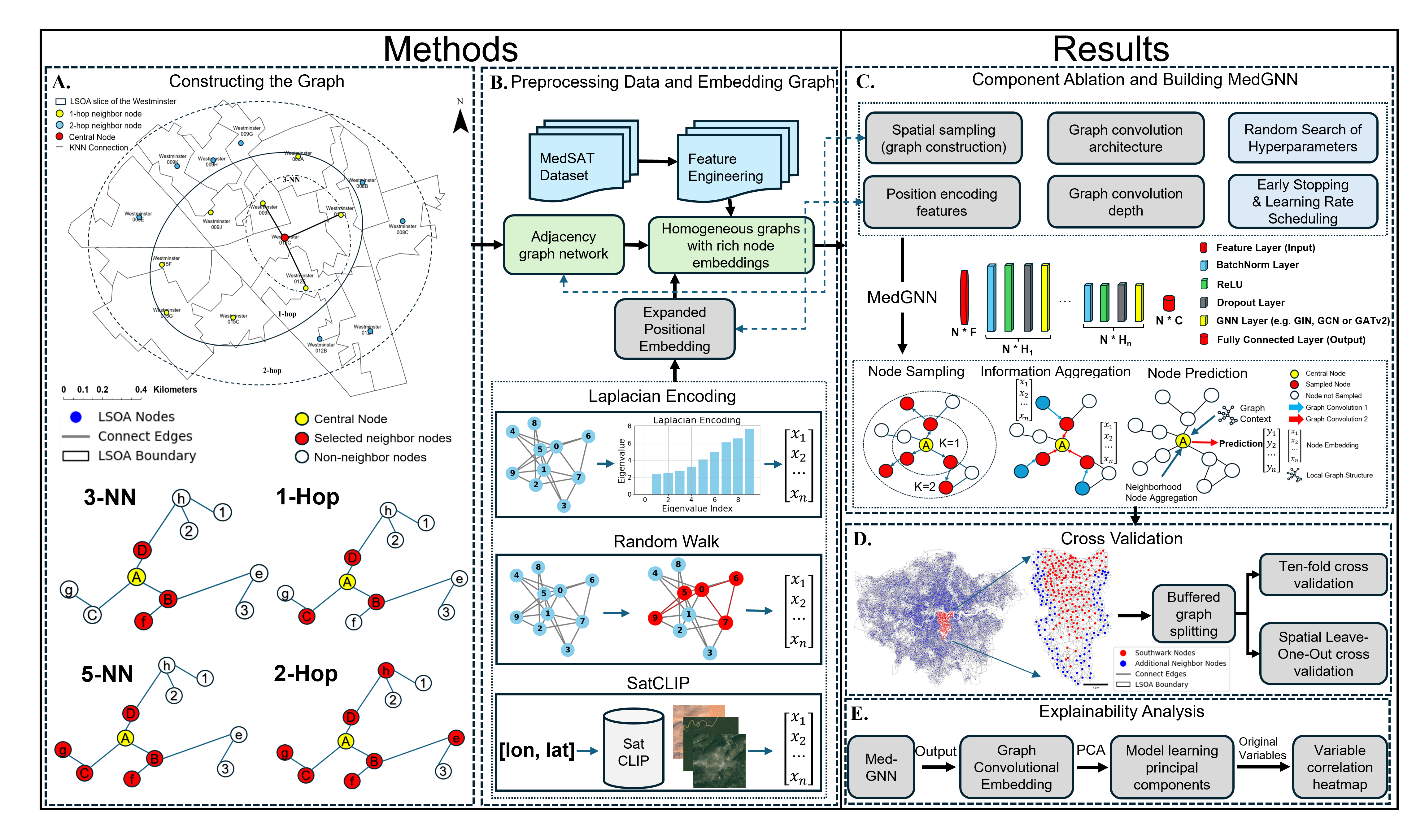} 
    \caption{UST-GNN methodology overview (Steps A--E).}
    \label{fig:overall_pipeline}
\end{figure}

Our methodology comprised five steps (Figure~\ref{fig:overall_pipeline}): (A) spatial graph construction, (B) data integration and positional/locational encoding, (C) GNN architecture development and ablation, (D) model evaluation, and (E) interpretability analysis. We describe each step below.

\subsection{Study Area}
\begin{figure}[!htbp]
    \centering
    \includegraphics[width=\textwidth]{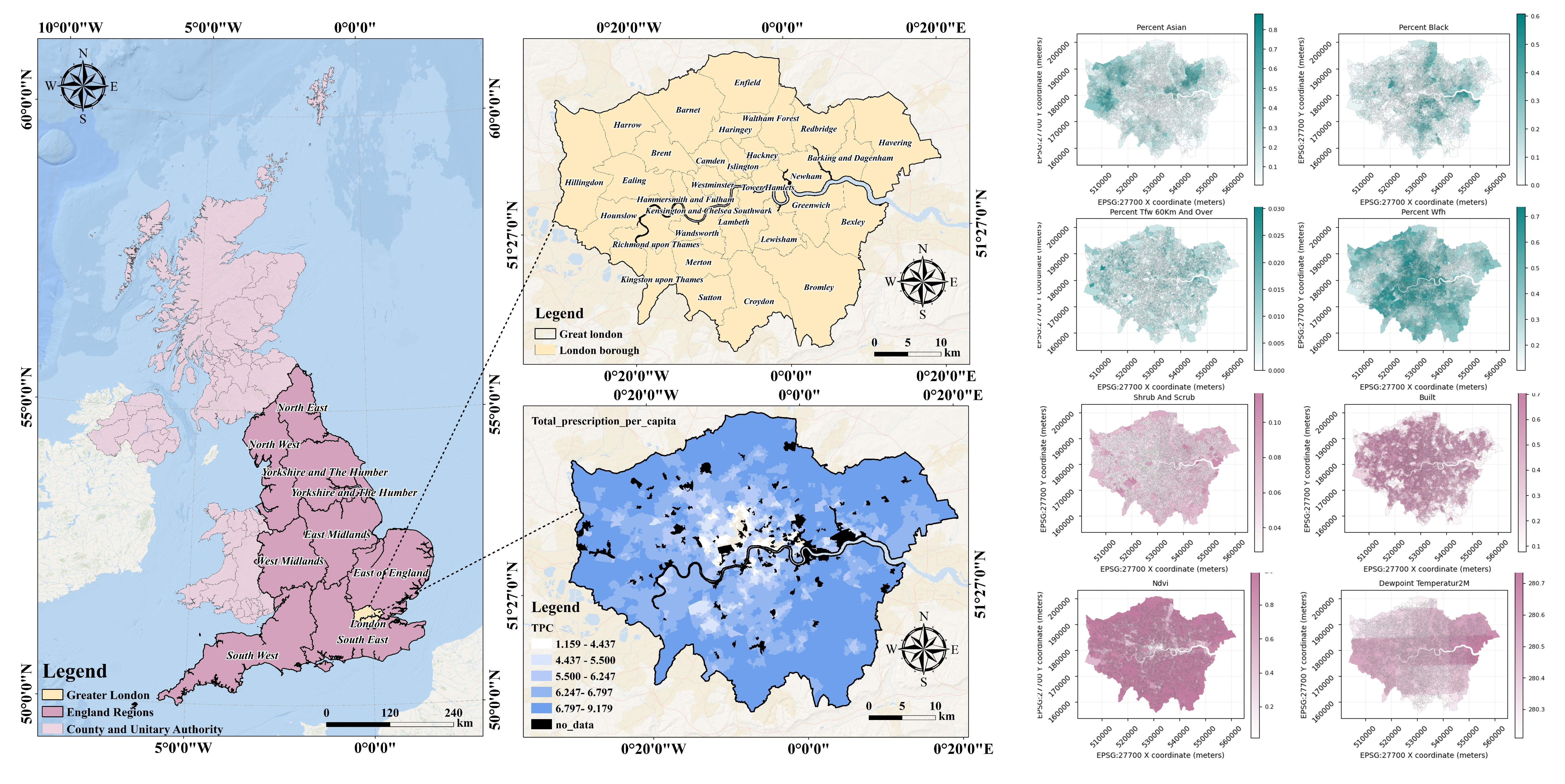}
    \caption{Study area (Greater London) and example spatial variables. Left: location within England. Centre: borough boundaries, LSOAs, and total per-capita prescription counts (TPC). Right: example socio-demographic and environmental predictors from the MedSAT dataset.}
    \label{fig:study_area}
\end{figure}

Our study focused on Greater London in the United Kingdom, comprising 32 London boroughs and the City of London (Figure~\ref{fig:study_area}). The analysis was conducted at the Lower Layer Super Output Area (LSOA) level, a small-area statistical geography in England, with each LSOA typically containing 1,000--3,000 residents and covering 1--4~\,km$^2$. Figure~\ref{fig:study_area} provides a geographic overview of the study area, the spatial distribution of total per-capita prescription counts (TPC), and example socio-demographic and environmental predictors from the MedSAT dataset, highlighting substantial spatial variation across Greater London.
\subsection{Step A: Constructing the Graph}\label{sec:stepA}
To cast the prediction of prescription outcomes in geographical neighborhoods as a node-level learning task, we represented the study area as a graph \( G=(V, E) \). Each LSOA corresponded to a node \( v_i \in V \), positioned at its geographic centroid. An edge \( e_{ij} \in E \) connected two nodes if the corresponding LSOAs are considered spatial neighbors. The adjacency relationships are encoded in a binary spatial proximity matrix \( A \in \mathbb{R}^{n \times n} \), where \( A_{ij}=1 \) if LSOA \( i \) and \( j \) are adjacent (and 0 otherwise). Each node \( v_i \) was associated with a feature vector \( \mathbf{x}_i \), comprising its socio-environmental attributes, and a target value \( y_i \), indicating the per capita prescription outcome of interest.

As shown in Part~A of Figure~\ref{fig:overall_pipeline}, multiple definitions of spatial adjacency, such as $k$-hop contiguity~\citep{chen2020iterative} and $k$-nearest neighbors~\citep{feng2022powerful}, were evaluated to determine the most informative graph topology. We addressed this selection process in Section~\ref{sec:method:UST-GNN ablation}, where an ablation study empirically evaluated the impact of adjacency construction methods.

\subsection{Step B: Preprocessing Data and Embedding Graph}
Having established the spatial graph structure in Step~A, the next step was to associate each node with a set of attributes representing the socio-environmental context of its corresponding geographic unit. This involved selecting, processing, and embedding multi-source urban indicators into the graph structure, ensuring that the feature representation was both comprehensive and robust. While our case study focused on neighbourhood-level health outcomes, the procedure described here is generalisable to other urban analytics tasks involving spatially structured data.

\subsubsection{MedSAT Dataset}
We utilised the MedSAT dataset~\citep{scepanovic2024medsat}, which integrates detailed socio-demographic and environmental indicators at the LSOA level. They are the smallest geographic units in England for which social and health indicators can be publicly released due to privacy constraints. MedSAT compiled data from multiple authoritative sources, including the Office for National Statistics (ONS)~\citep{ons2025}, the National Health Service (NHS)~\citep{nhs2025}, and remote sensing platforms such as Copernicus and Google Earth Engine (GEE)~\citep{gee2025}, ensuring consistency and accuracy across variables.

The dataset covered 111 socio-demographic indicators---such as gender ratio, age distribution, and measures of deprivation---and 55 environmental variables, including air quality metrics (e.g., NO\textsubscript{2}, PM\textsubscript{2.5}), greenness, land cover, and climate characteristics. All variables were aligned with the 2020 statistical and geographic boundaries of Greater London. In addition, MedSAT provided six per-capita medical prescription outcomes (anxiety, depression, diabetes, hypertension, asthma, and opioid use) derived from NHS administrative records. Although these outcomes are health-specific, the underlying multi-modal indicator set is structurally similar to datasets used in other domains such as environmental monitoring, mobility analysis, and socio-economic modelling, enabling broader applicability of the proposed framework.

\subsubsection{Feature selection and preprocessing}\label{sec:feature_selection}
Given the dataset's high dimensionality and potential multicollinearity, we performed systematic feature selection to enhance model stability, interpretability, and the attribution clarity of subsequent GNN analyses. The MedSAT dataset initially contained 166 variables (111 socio-demographic and 55 environmental indicators). 

Our feature selection process proceeded in two stages. First, we conducted pairwise correlation analysis to identify and remove highly redundant features ($|r|>0.90$), retaining within each correlated set the most representative variable informed by domain knowledge~\citep{lantz1998socioeconomic, apter1999influence, chen2008air, kjellstrom2007urban, nutsford2013ecological, marmot2005social}. Second, we applied additional dimensionality reduction techniques to further streamline the feature set while preserving theoretically essential predictors commonly used as key confounders in public health and socio-spatial research~\citep{bor2014child, harvey2017can, hossain2020epidemiology, hynie2018social, compton2015social}.

This process resulted in a refined feature set of 67 variables: 13 environmental indicators and 54 socio-demographic variables. The selected features encompass key demographic characteristics (age distribution, gender, ethnicity), socioeconomic indicators (occupation, housing, deprivation), and environmental exposures (air quality, greenness, land cover). Table~\ref{tab:control_variables} lists the key control and environmental variables retained across all configurations. Detailed information about the complete feature selection methodology and variable statistics is provided in the supplementary materials (~\ref{app:feature_selection}).

After filtering, all retained variables were normalised and embedded as node attributes $\mathbf{x}_i$ for each LSOA in the graph.

\subsubsection{Positional and Locational Encoding}\label{sec:encodings}
To supplement the socio-demographic and environmental attributes derived from MedSAT, we enriched each node with additional locational (spatial) and positional (topological) encodings. This design enabled the model to better capture both relative topological positions within the graph and absolute spatial context in the urban environment. Specifically, we incorporated three distinct encoding strategies---Laplacian positional encoding~\citep{dwivedi2021graph}, random walk encoding~\citep{schaub2020random}, and SatCLIP locational encoding~\citep{klemmer2023satclip}---to enhance spatial expressiveness. The locational encoder is implemented as a modular component that maps geographic context to a fixed-dimensional embedding. Alternative locational encoders, including other compatible geospatial foundation models, can therefore be substituted without requiring changes to the downstream UST-GNN pipeline. These were systematically evaluated in our ablation study (see Figure~\ref{fig:overall_pipeline}, Part~C), and implementation details are as follows:
\begin{table}[!htbp]
\centering
\footnotesize
\caption{Key control and environmental variables retained across all model configurations.}
\label{tab:control_variables}
\resizebox{\textwidth}{!}{%
\begin{tabular}{p{4.5cm} p{9.5cm}} 
\toprule
\textbf{Variable name}        & \textbf{Description}                                           \\ 
\midrule
\textbf{Demographics (Age)}  & Population proportion across age groups (e.g., Aged 10--14, 15--19, 20--24, \ldots, 85+); age structure is a known determinant of chronic disease prevalence and healthcare utilisation~\citep{lantz1998socioeconomic, marmot2005social}. \\
\textbf{Sex and ethnicity}   & Male share, Mixed-race share, White population share; demographic composition influences disease risk profiles and access to care~\citep{apter1999influence, marmot2005social}. \\
\textbf{Socioeconomic status} & Proportion in professional occupations, population density; socio-economic indicators are established predictors of health inequalities~\citep{lantz1998socioeconomic, marmot2005social, bixby2015associations}. \\
\textbf{Environment}         & NDVI (greenness), NO\textsubscript{2} concentration, water body, tree, grassland, and bare land coverage ratios; environmental exposures such as air quality and access to green space have demonstrated impacts on respiratory, cardiovascular, and mental health~\citep{chen2008air, kjellstrom2007urban, fecht2015associations, twohig2018health}. \\
\bottomrule
\end{tabular}%
}
\end{table}

\paragraph{Laplacian positional encoding}
Laplacian encoding computes node-level position features from the eigenvectors of the graph Laplacian matrix. This spectral technique captures the connectivity and smoothness of local graph neighbourhoods, assigning similar embeddings to structurally proximate nodes~\citep{dwivedi2021graph}. Let the graph Laplacian be defined as
\[
\mathbf{L} = \mathbf{D} - \mathbf{A},
\]
where $\mathbf{A}$ is the adjacency matrix and $\mathbf{D}$ is the degree matrix. We then extracted the first three non-trivial eigenvectors of $\mathbf{L}$ to represent each node’s relative position within the graph topology:
\[
\mathbf{L}\mathbf{u}_k = \lambda_k \mathbf{u}_k, \qquad
\mathbf{p}_i = \left[ u_1(i), u_2(i), u_3(i) \right].
\]
where $\lambda_k$ and $\mathbf{u}_k$ are the $k$-th eigenvalue and eigenvector of $\mathbf{L}$, respectively, and $\mathbf{p}_i$ is the 3-dimensional Laplacian positional encoding of node $i$, obtained from the entries of the first three non-trivial eigenvectors evaluated at node $i$.

\paragraph{Random walk positional encoding}
Random walk encoding represents each node by its probability distribution over a one-step traversal to its 1-hop neighbours~\citep{dwivedi2021graph, yeh2023random}. This approach reflects localized relational structures by modelling how information could propagate across directly connected areas. Given the moderate size and shallow depth of our spatial graph, a single-step walk was sufficient to capture local adjacency without overcomplicating the feature space:
\[
 P(v_i, t) = \frac{1}{Z} \sum_{j \in \mathcal{N}(v_i)} \frac{A_{ij}}{d_i} P(v_j, t-1)
 \]
where $P(v_i, t)$ is the probability distribution of node $v_i$ at time $t$, $d_i$ is the degree of $v_i$, and $Z$ is a normalization factor.

\paragraph{Geo-foundation-model-based locational encoding}
SatCLIP~\citep{klemmer2023satclip} provides absolute spatial descriptors by combining Sentinel-2 satellite imagery and geospatial coordinates into a unified 256-dimensional embedding, generated by a pretrained earth-observation (EO) contrastive learning GeoAI model. These embeddings encode physical and environmental characteristics from satellite images of each LSOA, such as vegetation, surface materials, and land use. To reduce dimensionality and ensure comparability with other node features, we applied PCA and retained the top principal components as the final locational encoding vector. To assess the modularity of this component, we also considered Google's AlphaEarth~\citep{brown2025alphaearth} as an alternative geo-foundation-model \emph{GeoFM} locational encoder in the ablation analysis. 

By appending these position and location embeddings to each node’s attribute vector $\mathbf{x}_i$, UST-GNN was equipped to learn from both relative topological patterns and absolute geographic contexts. This dual perspective is particularly valuable in urban health applications, where both network position (e.g., connectedness to high-risk areas) and absolute location (e.g., proximity to green space or pollution sources) can link to health outcomes. Section~\ref{sec:method:ablation} evaluates the individual and combined contributions of these encoding methods, including different position and \emph{GeoFM} locational encoders, through comparative experiments.

\subsection{Step C: Building UST-GNN via component ablation}
\label{sec:method:UST-GNN ablation}
With the spatial graph structure and enriched node attributes prepared in Steps~A and~B, we next designed a graph neural network architecture capable of leveraging both socio-environmental attribute and spatial topological information to predict neighbourhood-level outcomes. Our aim was to identify an architecture that could effectively capture the socio-environmental and spatial patterns most relevant to the case study on health outcomes, while remaining robust to alternative data contexts. To this end, we conducted an extensive component-wise evaluation, systematically varying graph definitions, feature encodings, and GNN architectures to determine the combination yielding the best predictive performance (Figure~\ref{fig:overall_pipeline}, Part~C).

\subsubsection{UST-GNN model architecture}
UST-GNN is a graph neural network that predicts the outcome for each LSOA node using socio-environmental attributes together with the spatial-topological relationships encoded in the graph. In the case study, the target outcome was the per-capita prescription rate. The model takes as input the MedSAT-derived socio-demographic and environmental features, augmented with positional and locational encodings, and propagates them through multiple graph neural network layers. At each layer, a node aggregates information from its neighbours (as defined by the adjacency $A$) and combines it with its current representation, enabling iterative propagation of relevant signals across the urban network. 

Through these stacked message-passing operations, node embeddings progressively integrate both local context and broader neighbourhood influences. After $L$ such layers, the resulting node embeddings encapsulate multi-scale structural patterns relevant to the prediction task. A final fully connected layer maps each node embedding to the target value, producing predictions for all LSOAs (Figure~1E). Formally, the model learns a function $f: \mathbf{x}_i \mapsto \hat{y}_i$ that minimises the mean squared error (MSE) between predicted and observed values:
\begin{equation}
\mathcal{L} = \frac{1}{N} \sum_{i=1}^{N} \left(y_i - \hat{y}_i\right)^2,
\end{equation}
where $N$ is the number of nodes, $y_i$ is the ground-truth outcome for node $i$, and $\hat{y}_i = f(\mathbf{x}_i)$ is the model prediction.

\subsubsection{Spatial Component Ablation Study}
\label{sec:method:ablation}

To determine the most effective design that considers spatial effects, we conducted a comprehensive sequential ablation study assessing the influence of key model components. All experiments were performed on the training set using 3-fold internal cross-validation to ensure stable validation performance. In addition, within each training fold, the data were further divided into a 70:20:10 ratio, where 10\% of the validation subset was reserved as a held-out early-stopping set to prevent overfitting and ensure convergence stability~\citep{meyes2019ablation,kedzie2018content}. The specific data splitting strategy for cross-validation, including how spatial adjacency relationships were preserved in train–validation partitions, is detailed later in Step~D: \textit{Spatial Cross-Validation and Baseline Comparison}. Hyperparameters were tuned via $10$ rounds of random search within a commonly used GNN hyperparameter range~\citep{adnan2022utilizing,zhou2022auto}, covering learning rate, hidden dimension, dropout rate, number of layers, and aggregation functions.

We systematically tested multiple options across three core components of the model. First, different definitions of the spatial adjacency graph were assessed (as described in Section~\ref{sec:stepA}) to capture meaningful spatial dependencies among neighborhoods. Second, the contribution of positional and locational encodings was evaluated by testing the inclusion of no encoding, single encodings, and combined encodings (detailed in Section~\ref{sec:encodings}). Third, we compared four widely used GNN architectures under different layer configurations: Graph Convolutional Network (GCN) \citep{kipf2016semi}, Graph Isomorphism Network (GIN) \citep{xu2018powerful}, GraphSAGE \citep{hamilton2017inductive}, and Graph Attention Network v2 (GATv2) \citep{brody2021attentive}~\citep{zhang2019graph,bhatti2023deep}. These architectures span a range of message-passing strategies, including linear aggregation (GCN), MLP-based transformations (GIN), neighbourhood sampling (GraphSAGE), and attention-based mechanisms (GATv2).

For each configuration, we recorded validation performance using mean squared error (MSE) and coefficient of determination ($R^2$) to identify the most effective combination of components and hyperparameter settings. Table~\ref{tab:gnn_hyperparams} summarises the hyperparameter ranges used in our random search, based on commonly adopted configurations in prior work~\citep{hamilton2017inductive, velivckovic2018deep, kipf2016semi, xu2018powerful, adnan2022utilizing, zhou2022auto}.

\subsection{Step D: Spatial Cross-Validation and Baseline Comparison}
\label{Cross-Validation and Baseline Comparison}
To rigorously evaluate UST-GNN’s performance, we compared its predictive accuracy against several baseline models under consistent spatial cross-validation settings. Following recent urban health modelling studies~\citep{scepanovic2024medsat,maitra2025location}, we selected a diverse set of baselines covering statistical, machine learning, geographically-enhanced, and graph-based approaches. 
The statistical baselines included the Spatial Lag Model (SLM)~\citep{anselin2009spatial} and Geographically Weighted Regression (GWR)~\citep{fotheringham2009geographically}, which explicitly incorporated spatial effects within global and local linear frameworks, respectively. The machine learning baselines consisted of Random Forest and LightGBM~\citep{ke2017lightgbm}, representing non-spatial, non-linear ensemble learners. 
To further examine the benefit of spatial feature augmentation, we implemented geographically-enhanced variants of tree-based models: Geo-LightGBM and Geo-XGBOOST using two strategies: (i) directly providing geographic coordinates as additional input features, and (ii) incorporating a spatial weight matrix to encode neighborhood structure~\citep{li2022extracting,nikparvar2021machine}.

\begin{table}[!htbp]
\centering
\footnotesize
\caption{GNN hyperparameter search space for random search.}
\label{tab:gnn_hyperparams}
\begin{tabular}{p{4cm} p{8cm}}
\toprule
\textbf{Hyperparameter} & \textbf{Search range} \\
\midrule
Learning rate & \(\{1\!\times\!10^{-4},\ 5\!\times\!10^{-4},\ 1\!\times\!10^{-3},\ 5\!\times\!10^{-3}\}\) \\
Hidden dimension & \(\{16,\ 32,\ 64,\ 128,\ 256\}\) \\
Dropout rate & \(\{0.0,\ 0.2,\ 0.3\}\) \\
Aggregation function & \{\texttt{mean},\ \texttt{sum},\ \texttt{max}\} \\
Activation function & \{\texttt{ReLU},\ \texttt{ELU},\ \texttt{LeakyReLU}\} \\
Weight decay & \(\{0,\ 1\!\times\!10^{-5},\ 5\!\times\!10^{-5},\ 1\!\times\!10^{-4},\ 5\!\times\!10^{-4}\}\) \\
Batch size & \(\{32,\ 64,\ 128\}\) \\
Optimizer & \{\texttt{Adam},\ \texttt{AdamW}\} \\
\bottomrule
\end{tabular}
\end{table}
Finally, representative GNN baselines (GCN, GraphSAGE, and GIN) were also evaluated in the ablation study for comparison with UST-GNN’s spatial-topological architecture. For a fair comparison, all baseline models were trained and evaluated using the same 3-fold data splits and identical feature sets as UST-GNN. In practice, the statistical models (SLM and GWR) were fit on the training portion of each fold and applied directly to the test fold, whereas the machine learning and geographically-enhanced models were tuned within the training folds using internal validation splits to ensure optimal hyperparameter selection and stable performance.

\begin{figure}[!htbp]
    \centering
    \includegraphics[width=\textwidth]{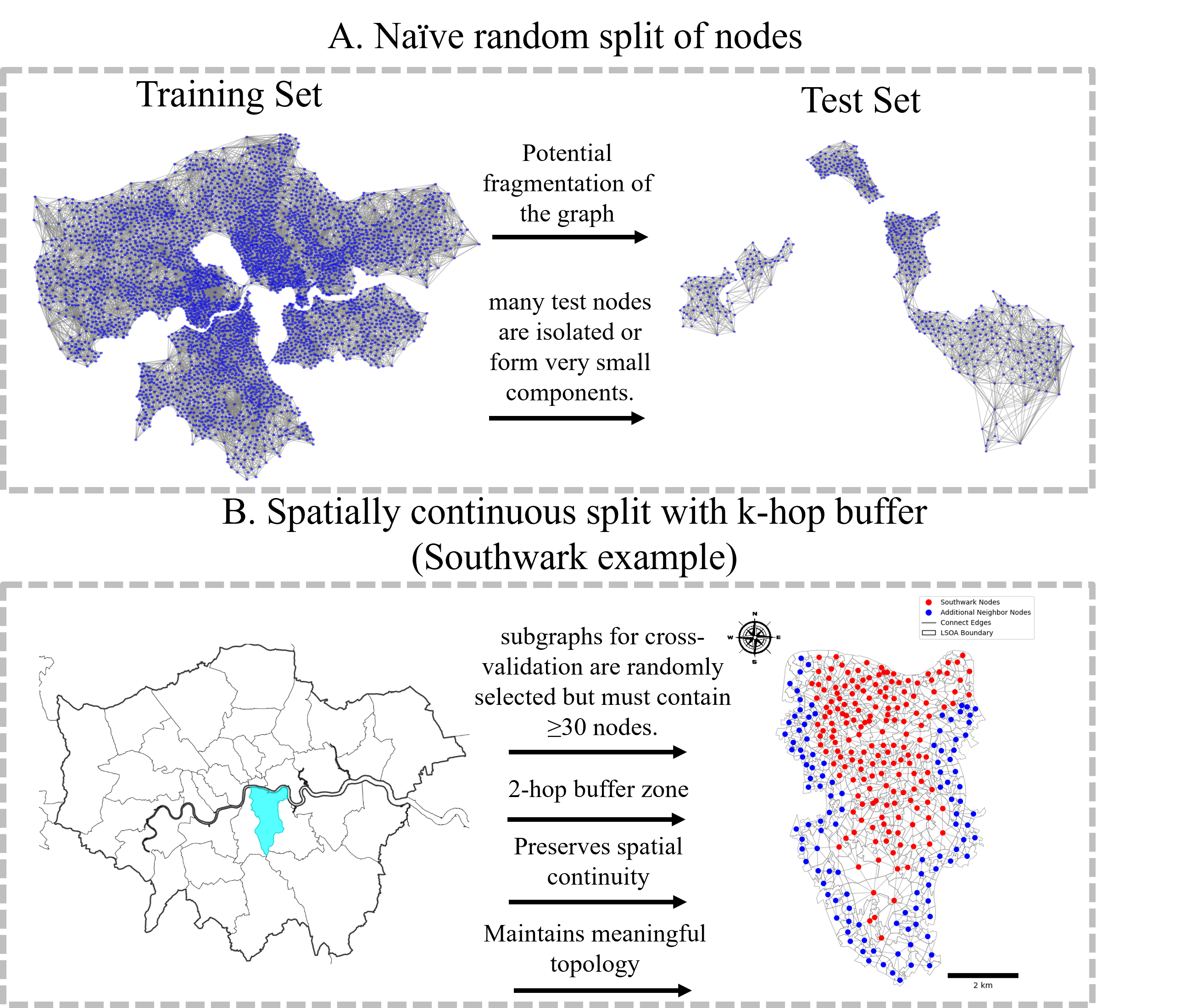}
    \caption{Comparison of splitting strategies for spatial graph cross-validation: (A) naïve random split; (B) spatially continuous split with a $k$-hop buffer.}
    \label{fig:split}
\end{figure}

\paragraph{Splitting strategies and spatial considerations}
Random node-level splits can severely fragment the spatial graph, often producing isolated test nodes or very small disconnected components. Spatial continuity is therefore critical for meaningful cross-validation. Figure~\ref{fig:split} compares a naïve random split (Panel~A) with our spatially continuous, buffer-based strategy (Panel~B). In the main evaluation setting, this strategy uses a fixed 2-hop buffer, while in the topology ablation study the exclusion region is scaled with the tested adjacency definition to ensure a fairer comparison across graph constructions with different receptive fields.

In our approach, subgraphs for cross-validation were randomly selected but constrained to contain at least 30 nodes, ensuring that each test region had sufficient spatial extent and internal connectivity. We further adopted a community-based graph partitioning scheme (using Louvain or METIS clustering, depending on graph size~\citep{shao2015community}) to generate spatially coherent folds, thereby preserving spatial coherence and graph connectivity across folds.

To mitigate spatial information leakage, we introduced a $k$-hop buffer around the test subgraph. In the main evaluation setting, we used a fixed 2-hop buffer, excluding this region from both training and testing so that no node used for prediction had seen direct or second-order neighbours during training. For the topology ablation study, however, we further adapted this protocol so that the exclusion region scales with the tested adjacency definition: for an $n$-hop graph, the held-out centre nodes are separated from training visibility by an $n{+}2$-hop invisible neighbourhood. This topology-aware exclusion rule is intended to reduce boundary-effect asymmetry when comparing graph constructions with different effective receptive fields. Conceptually, this corresponds to an inductive evaluation setting: model parameters are learned from one part of the city (training subgraphs) and tested on spatially disjoint regions, rather than reusing the same adjacency structure as in a transductive setup. This design better approximates real-world deployment scenarios, such as predicting health outcomes in unseen districts while preserving the contextual dependencies necessary for spatial reasoning.
\subsubsection{Ten-Fold Spatial Cross-Validation}
In the first strategy, we applied standard $k$-fold cross-validation with $k=10$~\citep{fushiki2011estimation}, incorporating the buffered splitting procedure described above. In each fold, the test set and its immediate 2-hop neighbors were withheld from training, minimizing border effects and preserving spatial structure. Performance was averaged over the 10 folds, providing a robust estimate of out-of-sample accuracy across the city while utilizing 90\% of the data in each training round.

\subsubsection{Leave-One-Borough-Out Spatial Cross-Validation}
The second strategy was a geographically structured leave-one-out cross-validation~\citep{wong2015performance}. In contrast to ten-fold CV, which partitions the data into multiple smaller folds, this approach held out an entire London borough during each evaluation. Six diverse boroughs: Camden, Barking and Dagenham, Barnet, Southwark, Croydon, and Ealing, these were selected to represent various geographic locations and socio-demographic profiles~\citep{raco2014urban}. The splitting procedure followed the same 2-hop buffered approach as in ten-fold CV to maintain spatial continuity.

This strategy served as a stringent test of spatial generalizability, revealing how well UST-GNN performed in unseen regions. Both cross-validation setups were evaluated using root mean squared error (RMSE), mean absolute error (MAE), and coefficient of determination ($R^2$). Hyperparameters were fixed across all folds, based on prior tuning through random search on each of the specific training sets.

\subsection{Step E: Interpretability analysis}
\begin{figure}[htbp]
    \centering
    \includegraphics[width=1\textwidth]{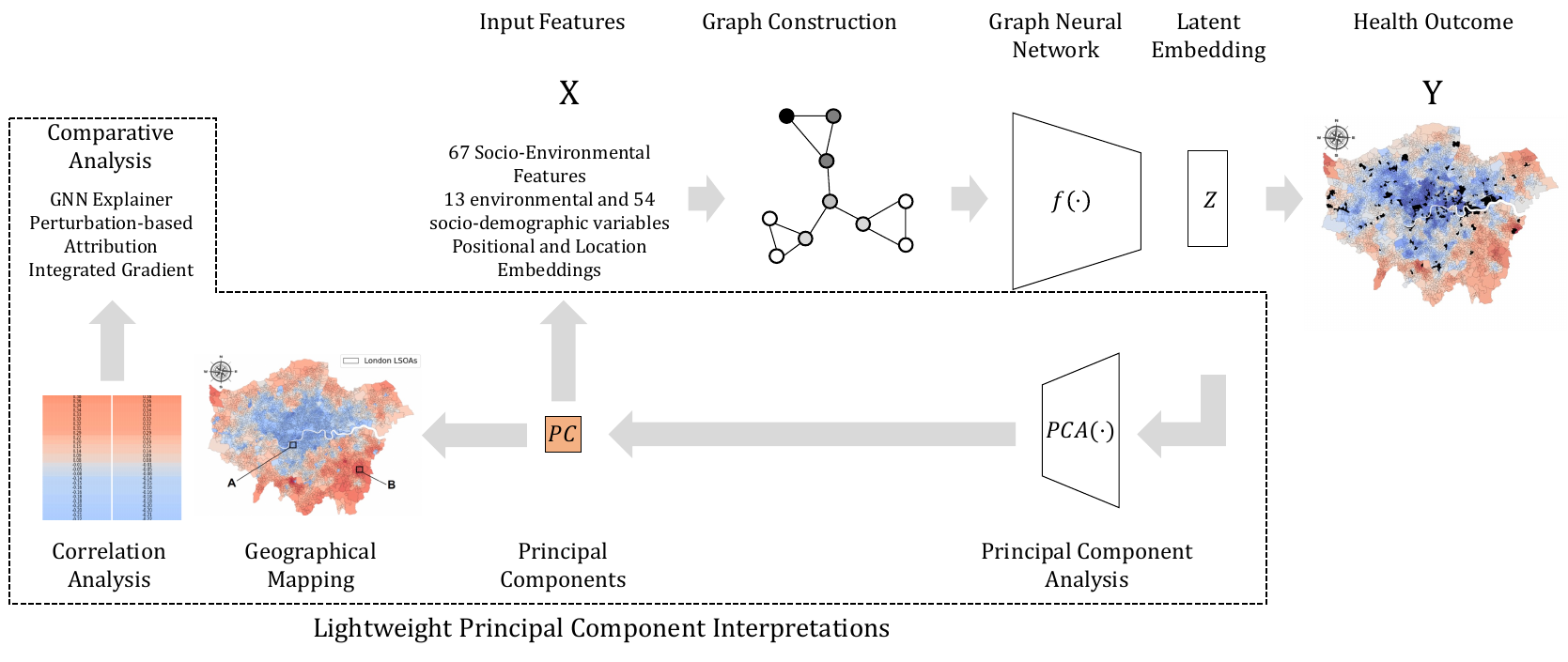}
    \caption{Principal-component-based interpretability framework and comparison with existing explainability methods.}
    \label{fig:PCI}
\end{figure}

To complement UST-GNN’s predictive capacity with meaningful interpretation, we introduced a light-weight principal component interpretation framework that globally interpret the input features through geographical mapping and correlational analysis as shown in Figure~\ref{fig:PCI}. The approach first applies PCA~\citep{greenacre2022principal, cohen2013applied} to the final-layer node embeddings to extract two intrinsic components that summarises the latent spatial structures and attributes learned by the model. 

The top two principal components (explaining over 80\% of variance) were then mapped back to neighbourhoods, enabling interpretation of embeddings as coherent spatial patterns rather than abstract features. The two principal components were then correlated with the original socio-demographic and environmental variables, thereby anchoring the intrinsic representation space back to real-world policy-relevant urban characteristics and revealing how the model internally organises outcome-relevant information. This approach provides a simplified association between the learned latent space and its policy-relevant input variables.

Unlike input-level explainers such as GNNExplainer or gradient-based methods, this global approach offers a holistic and distilled view of how message passing encodes spatial and socio-environmental dependencies. PCA provides a light-weight and interpretable means of identifying the dominant and instrinsic spatial hierarchies captured during learning, without relying on model-specific assumptions or task labels.

To further contextualise this design, we established a comparative protocol (~\ref{app:explainability}) against established explanation methods, Permutation-based attribution, Integrated Gradients, and GNNExplainer, under identical data splits and hyperparameters. We evaluated (i) consistency of variable importance, (ii) ability to produce spatially continuous gradients, and (iii) interpretability across local and global levels. This positions our PCA-based framework within a unified methodological benchmark for spatial model interpretation.

\section{Results}
In this section, we evaluate the UST-GNN framework through a series of experiments designed to: \emph{(1)} identify optimal model components via systematic ablation (Section~\ref{sec:res_ablation}), \emph{(2)} assess predictive performance in a health-related case study using the MEDSAT dataset (Section~\ref{sec:res_performance}), and \emph{(3)} illustrate interpretability insights from a domain-specific analysis of antidepressant prescriptions (Section~\ref{sec:res_XAI}). While the case study focuses on health outcomes, the experimental design and evaluation procedures are general to other urban analytics tasks involving spatially structured data.

\subsection{Component Ablation}
\label{sec:res_ablation}
To determine the optimal design of the UST-GNN framework, we performed a sequential ablation study covering four core components: message passing architecture, spatial adjacency definition, network depth, and location/positional encoding strategy. 

\begin{figure}[htbp]
    \centering
    \includegraphics[width=1\textwidth]{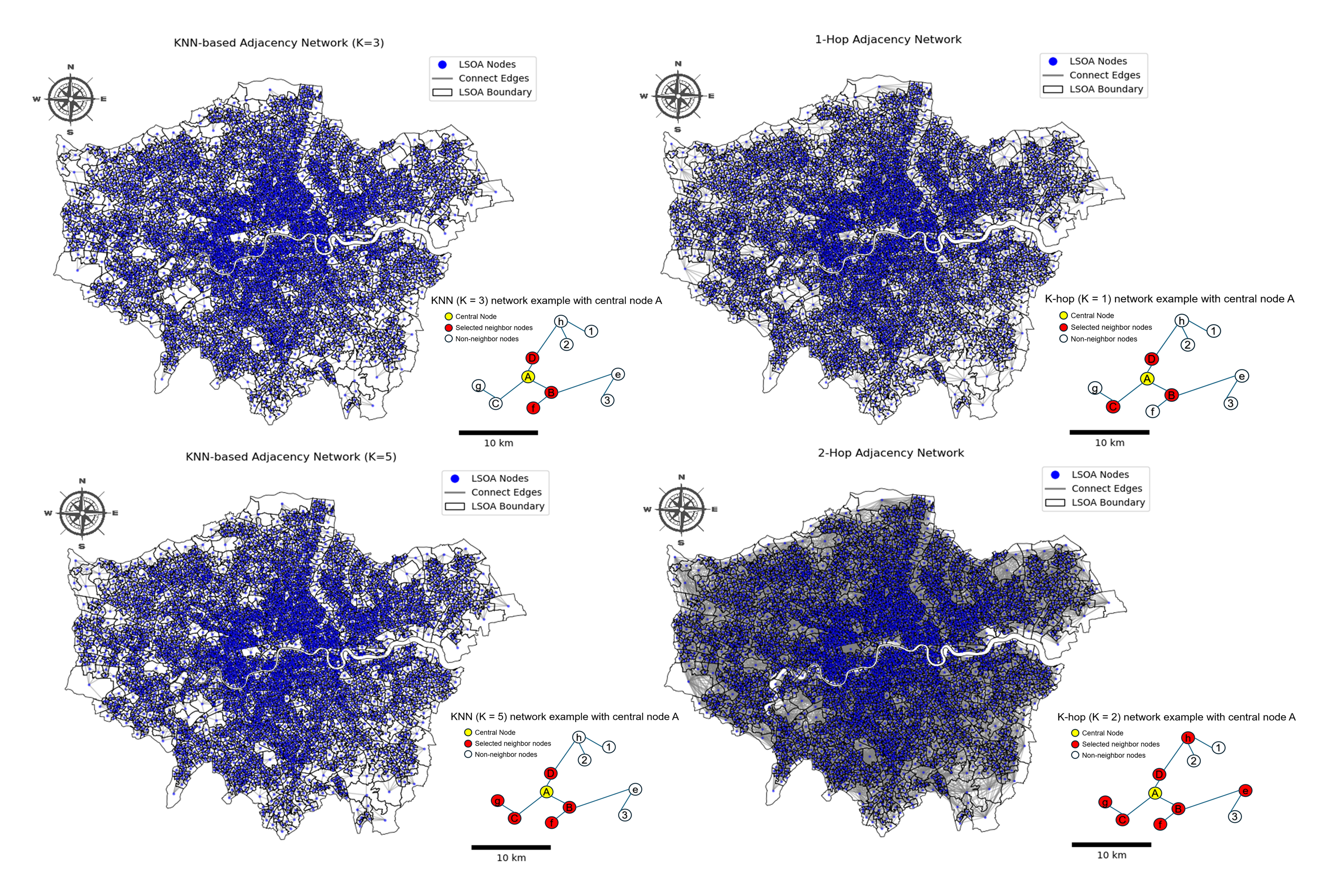}
    \caption{Comparison of spatial adjacency construction strategies for London LSOAs: KNN graphs ($k=3,5$) and $h$-hop graphs ($h=1,2$).}
    \label{fig:lsoa_adjacency}
\end{figure}

Figure~\ref{fig:lsoa_adjacency} illustrates the spatial adjacency construction strategies considered in the ablation study. KNN graphs generate relatively sparse and distance-based connections, whereas $h$-hop contiguity graphs produce progressively denser and more interconnected neighbourhood networks, directly affecting information flow, spatial scale, and computational cost in graph learning.

Table~\ref{tab:ablation_all} summarises the sequential ablation procedure, in which the best-performing component at each stage is retained and fixed for the next stage. The selected backbone inherited by the later stages is 2-head GATv2 with 1-HOP adjacency and 2 layers; the subsequent position-and-location-encoding blocks further refine this backbone, including comparisons across alternative encoders.

Among the adjacency configurations, the 1-HOP contiguity graph achieved the strongest cross-validated performance ($R^2=0.860 \pm 0.008$), outperforming both KNN alternatives and denser multi-hop graphs. In particular, extending the graph to 2-HOP and 3-HOP reduced performance to $0.821 \pm 0.027$ and $0.799 \pm 0.011$, respectively. This suggests that, for neighbourhood-level prescription prediction in Greater London, immediate contiguity captures the most relevant local spatial context, whereas denser topologies may introduce less informative meso-scale structure and increase the risk of over-smoothing. We therefore retain 1-HOP adjacency as the default spatial representation for UST-GNN.

The Graph Attention Network v2 (GATv2) emerged as the most effective architecture. Two graph convolutional layers provided the best balance between expressiveness and stability, with three layers remaining competitive but slightly less robust, and the additional head ablation showed that two attention heads performed best overall. For position encoding, the combination of SatCLIP and Random Walk yielded the highest performance, suggesting that both absolute geographic positioning and local structural context contribute to stronger spatial representations. The additional AlphaEarth comparison further shows that different \emph{GeoFM}-based representation enhancements provide broadly similar improvements within the same UST-GNN pipeline, indicating that the framework can benefit from multiple interchangeable locational encoding modules. Together, these design choices define the final UST-GNN configuration used in the subsequent experiments.

\subsection{Model Performance}\label{sec:res_performance}
Addressing our second contribution, we evaluated UST-GNN in a 10-fold as well as leave-one-out cross-validation.

\subsubsection{Ten-Fold Spatial Cross-Validation}
\label{sec:res_CV}
We evaluated UST-GNN against a comprehensive suite of baseline models, including statistical methods (Spatial Lag Model, SLM; Geographically Weighted Regression, GWR), machine learning approaches (Random Forest, LightGBM), geographically-enhanced models (Geo-XGBOOST and Geo-LightGBM incorporating either coordinate or spatial-weight features), and graph neural network baselines (GCN, GraphSAGE) that share the same adjacency structure as UST-GNN. All models were trained and tested under an identical 10-fold cross-validation protocol to ensure comparability.

\begin{table}[htbp]
\centering
\caption{Component-wise ablation results for depression prediction. The ablation study was conducted sequentially starting from a preset backbone configuration (GCN; 1-hop adjacency; 2 layers; 2 attention heads; no positional/location embeddings). At each step, one component was varied across candidate settings, previously selected components were fixed to their selected values, and all remaining untested components retained their preset settings. The final selected model is reported in the last row.}
\label{tab:ablation_all}
\scriptsize
\setlength{\tabcolsep}{4pt}
\renewcommand{\arraystretch}{0.98}
\resizebox{\columnwidth}{!}{%
\begin{tabular}{lcl}
\toprule
\textbf{Tested Pluggable Module} & \textbf{Setting / Candidate} & \textbf{3-fold $R^2$} \\
\midrule

\multirow{6}{*}{\textbf{Architecture}}
& GCN & 0.805 {\tiny$\pm$ 0.013} \\
& GIN & 0.811 {\tiny$\pm$ 0.006} \\
& GraphSAGE & 0.828 {\tiny$\pm$ 0.009} \\
& GAT & 0.829 {\tiny$\pm$ 0.013} \\
& \textbf{GATv2$^*$} & 0.860 {\tiny$\pm$ 0.008} \\
& \textit{Best architecture updated to GATv2.} & \\
\midrule

\multirow{6}{*}{\textbf{Adjacency}}
& 3-NN & 0.837 {\tiny$\pm$ 0.009} \\
& 5-NN & 0.854 {\tiny$\pm$ 0.006} \\
& \textbf{1-hop$^*$} & 0.860 {\tiny$\pm$ 0.008} \\
& 2-hop & 0.821 {\tiny$\pm$ 0.027} \\
& 3-hop & 0.799 {\tiny$\pm$ 0.011} \\
& \textit{Best adjacency = 1-hop.} & \\
\midrule

\multirow{5}{*}{\textbf{Convolution Depth}}
& 1 layer & 0.767 {\tiny$\pm$ 0.014} \\
& \textbf{2 layers$^*$} & 0.860 {\tiny$\pm$ 0.008} \\
& 3 layers & 0.858 {\tiny$\pm$ 0.011} \\
& 4 layers & 0.833 {\tiny$\pm$ 0.009} \\
& \textit{Best depth = 2 layers.} & \\
\midrule

\multirow{5}{*}{\textbf{Attention Head}}
& 1 head & 0.840 {\tiny$\pm$ 0.011} \\
& \textbf{2 heads$^*$} & 0.860 {\tiny$\pm$ 0.008} \\
& 4 heads & 0.853 {\tiny$\pm$ 0.004} \\
& 8 heads & 0.853 {\tiny$\pm$ 0.013} \\
& \textit{Best number of attention heads = 2.} & \\
\midrule

\multirow{10}{*}{\textbf{Position/Location}}
& None & 0.860 {\tiny$\pm$ 0.008} \\
& Random Walk (1-D)$^*$ & 0.861 {\tiny$\pm$ 0.008} \\
& Laplace (3-D) & 0.854 {\tiny$\pm$ 0.011} \\
& SatCLIP (2-D) & 0.855 {\tiny$\pm$ 0.008} \\
& AlphaEarth (2-D) & 0.861 {\tiny$\pm$ 0.008} \\
& AlphaEarth + Laplace (2+3-D) & 0.857 {\tiny$\pm$ 0.018} \\
& AlphaEarth + Random Walk (2+2-D) & 0.863 {\tiny$\pm$ 0.011} \\
& SatCLIP + Laplace (2+3-D) & 0.859 {\tiny$\pm$ 0.006} \\
& \textbf{SatCLIP + Random Walk (2+2-D)$^*$} & \textbf{0.868 {\tiny$\pm$ 0.008}} \\
& \textit{Best positional/location encoding updated to SatCLIP + Random Walk.} & \\
\midrule

\textbf{Final model}
& \textbf{UST-GNN = GATv2; 1-hop; 2 layers; 2 heads; SatCLIP + Random Walk}
& \textbf{0.868 {\tiny$\pm$ 0.008}} \\
\bottomrule
\end{tabular}%
}
\end{table}

\begin{figure}[!h]
  \centering
  \includegraphics[width=\textwidth]{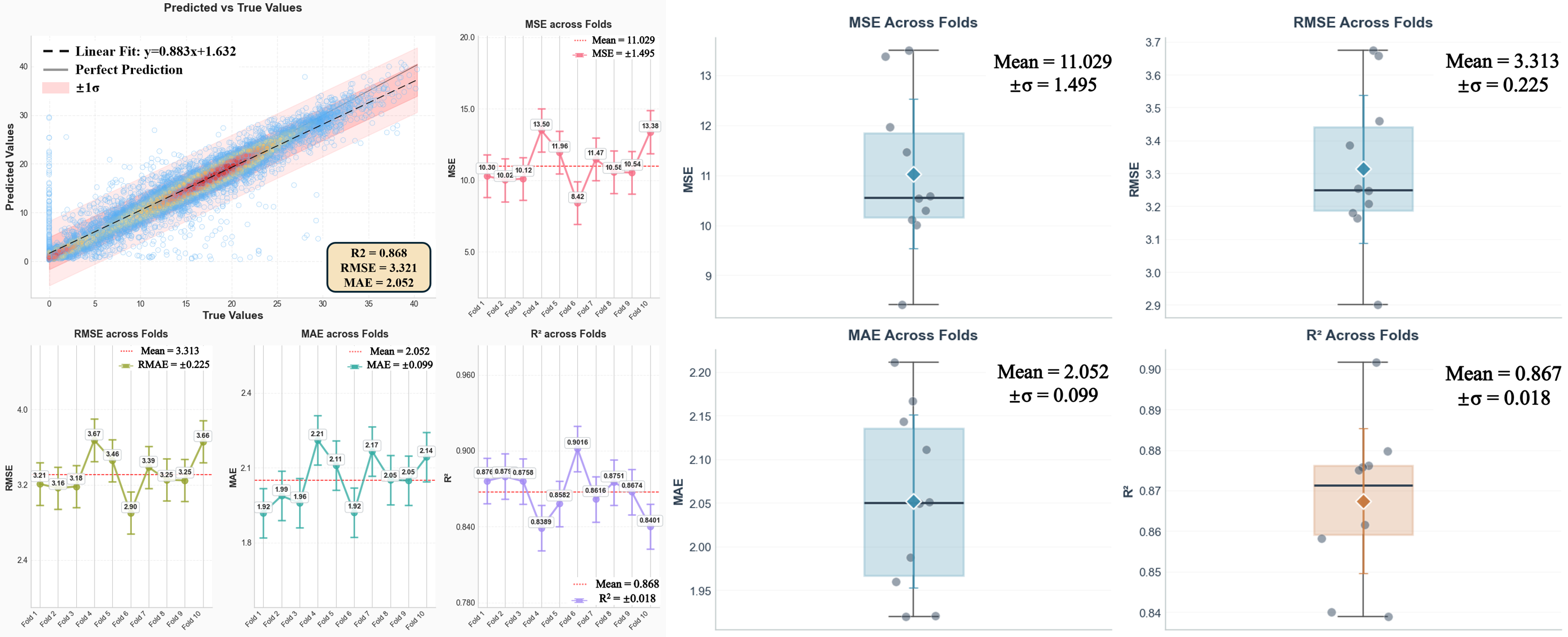}
  \caption{Predictive performance and cross-validation stability of UST-GNN for depression prescriptions: predicted vs.\ observed values, fold-wise metrics, and metric distributions across folds.}
  \label{fig:depression_performance}
\end{figure}

Table~\ref{tab:baseline_comparison} summarises predictive performance across all six prescription outcomes, while Figure~\ref{fig:depression_performance} provides a detailed view of fold-wise performance and prediction stability for the representative depression task. UST-GNN consistently outperformed all baselines across diabetes, hypertension, asthma, depression, anxiety, and opioid prescriptions. Best and second-best results are highlighted in bold and underline, respectively. While geographically enhanced models achieved notable gains over traditional machine learning baselines, confirming the value of incorporating spatial context, UST-GNN further improved predictive accuracy by explicitly modelling graph-based spatial dependencies and nonlinear socio-environmental interactions. Across outcomes, it improves upon the strongest baseline by 8.4\%--13.2\% in out-of-sample $R^2$. Figure~\ref{fig:depression_performance} further shows that these gains are accompanied by accurate and stable out-of-fold predictions with limited variation across spatial folds.

\begin{table*}[!htbp]
  \caption{Ten-fold cross-validation results across six prescription outcomes. Best results are in bold, second-best are underlined.}
  \label{tab:baseline_comparison}
  \centering
  \resizebox{\textwidth}{!}{%
    \begin{tabular}{lllllll}
      \toprule
      \textbf{Model} & \textbf{Diabetes} & \textbf{Hypertension} & \textbf{Asthma} & \textbf{Depression} & \textbf{Anxiety} & \textbf{Opioids} \\
      \midrule
      \textbf{Statistical Baselines} & & & & & & \\
      SLM & 0.711 & 0.725 & 0.684 & 0.714 & 0.678 & 0.677 \\
      GWR & 0.726 & 0.752 & 0.710 & 0.734 & 0.705 & 0.691 \\
      \midrule
      \textbf{Machine Learning Baselines} & & & & & & \\
      Random Forest & 0.637 {\tiny ± 0.021} &  0.671 {\tiny ± 0.016} & 0.621 {\tiny ± 0.021} &  0.633 {\tiny ± 0.023} & 0.599 {\tiny ± 0.023} & 0.555 {\tiny ± 0.022} \\
      LightGBM & 0.632 {\tiny ± 0.029} & 0.649 {\tiny ± 0.017} & 0.589 {\tiny ± 0.024} & 0.628 {\tiny ± 0.020} & 0.586 {\tiny ± 0.021} & 0.535 {\tiny ± 0.032} \\
      \midrule
      \textbf{Geographically-Enhanced Models} & & & & & & \\
      Geo-XGBOOST(Coordinates) & 0.691 {\tiny ± 0.023}& 0.725 {\tiny ± 0.013} & 0.678 {\tiny ± 0.016} & 0.699 {\tiny ± 0.019} & 0.662 {\tiny ± 0.017} & 0.6546 {\tiny ± 0.021} \\
      Geo-LightGBM (Coordinates) & 0.684 {\tiny ± 0.022} & 0.720 {\tiny ± 0.016} & 0.670 {\tiny ± 0.021} & 0.689 {\tiny ±0.017} & 0.655 {\tiny ± 0.021} & 0.627 {\tiny ± 0.020} \\
      Geo-XGBOOST(Spatial weight matrix) & \underline{0.781 {\tiny ± 0.023}} & \underline{0.801 {\tiny ± 0.017}} & \underline{0.764 {\tiny ± 0.022}} & 0.767 {\tiny ± 0.022} & \underline{0.762 {\tiny ± 0.021}} & \underline{0.751 {\tiny ± 0.015}} \\
      Geo-LightGBM (Spatial weight matrix) & 0.772 {\tiny ± 0.026}  & 0.790 {\tiny ± 0.019} & 0.754 {\tiny ± 0.023} & \underline{0.779 {\tiny ± 0.015}} & 0.750 {\tiny ± 0.020} & 0.735 {\tiny ± 0.013} \\
      \midrule
      \textbf{Graph Neural Network Baselines} & \multicolumn{6}{l}{Evaluated (GCN, GraphSAGE, GIN) in ablation; see Tab.~\ref{tab:ablation_all}} \\
      \midrule
      \textbf{Proposed Method} & & & & & & \\
      UST-GNN (Ours) & \textbf{0.864 {\tiny ± 0.024}} & \textbf{0.869 {\tiny ± 0.020}} & \textbf{0.849 {\tiny ± 0.018}} & \textbf{0.868 {\tiny ± 0.018}} & \textbf{0.853 {\tiny ± 0.021}} & \textbf{0.850 {\tiny ± 0.025}} \\
      \midrule
      {Relative gain over the best baseline (R²)} & 10.6\% & 8.4\% & 11.1\% & 11.4\% & 11.9\% & 13.2\% \\
      \bottomrule
    \end{tabular}
  }
\end{table*}

\begin{figure}[!htbp]
    \centering
    \includegraphics[width=0.96\textwidth]{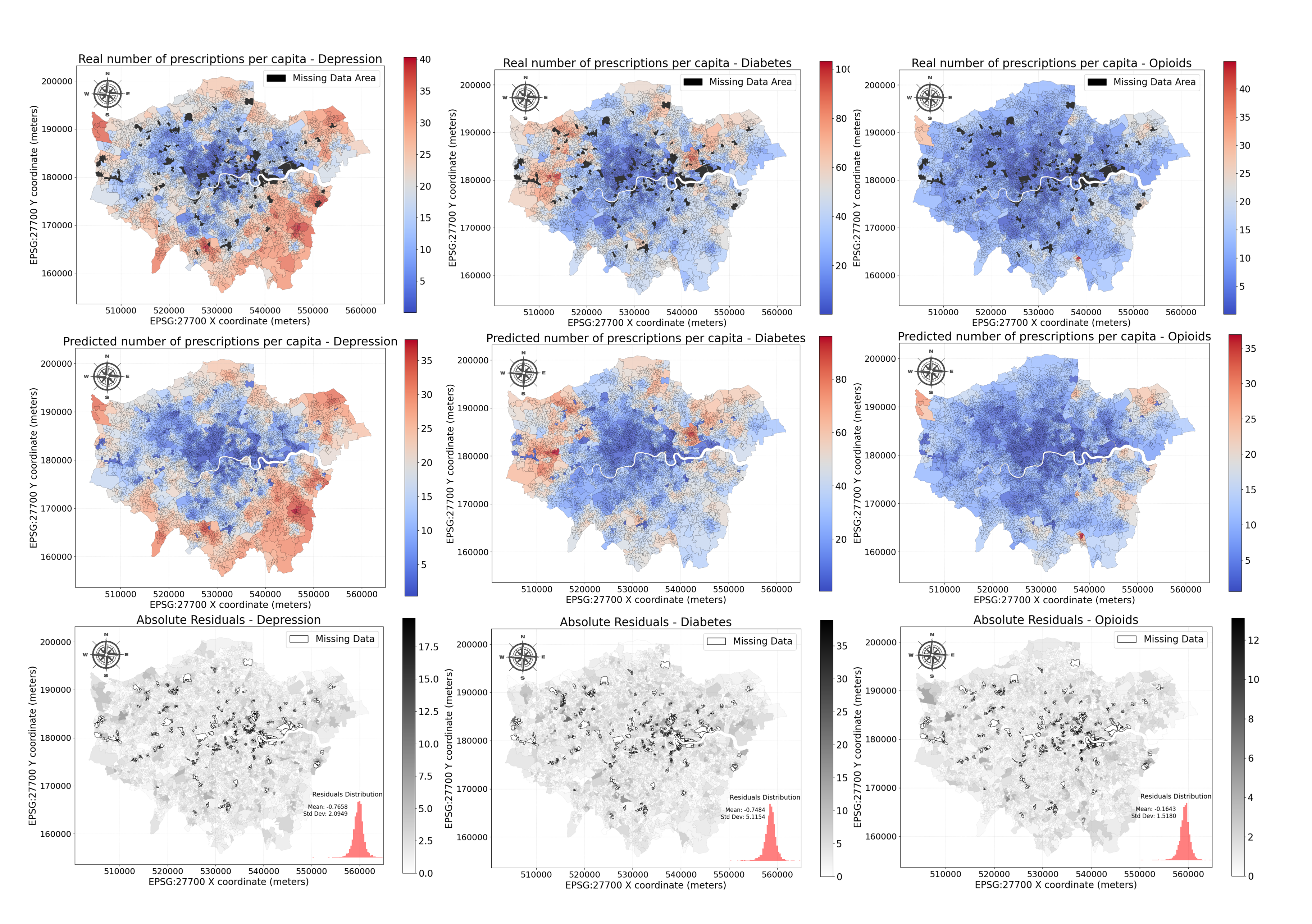}
    \caption{
    Spatial distribution of real values, predicted values, and absolute residuals for three representative prescription categories in London: 
    (left) Depression (mental health), 
    (middle) Diabetes (chronic disease), and 
    (right) Opioids (pain management). 
    Rows correspond to: 
    (top) real number of prescriptions per capita, 
    (middle) predicted number of prescriptions per capita by the UST-GNN framework, and 
    (bottom) absolute residuals between predictions and real values. 
    Blue-to-red gradients indicate relative prescription intensity, 
    while grayscale in the bottom row shows residual magnitude (darker = larger error). 
    Black polygons mark LSOAs with missing target values; histograms in the residual maps show the residual distribution with mean and standard deviation.
    }
    \label{fig:outcome}
\end{figure}

To further illustrate predictive performance, we visualised observed versus predicted per-capita prescription rates across three representative health outcomes: depression (mental health), diabetes (chronic disease), and opioids (pain management) in Figure~\ref{fig:outcome}. For each outcome, the UST-GNN configuration was individually fine-tuned. Across outcomes, the model successfully reproduced major spatial trends—such as high depression prescription rates in South East London, elevated diabetes prescriptions in eastern boroughs, and opioid hotspots in outer boroughs—while capturing low-prescription clusters in Central London. Residual maps indicate generally small errors, with localized discrepancies in parts of Central and East London, suggesting potential effects of unobserved socio-environmental factors or limitations in current graph connectivity definitions.

\subsubsection{Leave-One-Out Spatial Cross-Validation (LOOSCV)}
To further test spatial generalisability under a more stringent setting, We conducted LOOSCV using six Boroughs in London---Camden, Barking, Barnet, Southwark, Croydon, and Ealing, which were selected for their diverse geographic, socio-demographic, and environmental characteristics~\citep{ng1997preventing,kearns1997algorithmic}. Table~\ref{tab:loocv_results} summarized the results across six health outcomes.

\begin{table}[!htbp]
  \caption{Performance of UST-GNN under Leave-One-Out Spatial Cross-Validation (LOOSCV) across six London boroughs and six prescription outcomes. Best $R^2$ values per medical outcome are shown in bold.}
  \label{tab:loocv_results}
  \centering
  \footnotesize
  \setlength{\tabcolsep}{3pt}
  \renewcommand{\arraystretch}{0.45}
  \begin{tabular}{llcccccc}
    \toprule
    \textbf{Outcome} & \textbf{Metric} & Camden & Barking & Barnet & Southwark & Croydon & Ealing \\
    \midrule

    \multirow{3}{*}{Diabetes} 
    & RMSE & 3.93 & 7.03 & 6.83 & 5.61 & 4.60 & 10.43 \\
    & MAE  & 2.80 & 5.28 & 4.84 & 4.11 & 3.34 & 6.80 \\
    & $R^2$ & 0.78 & 0.88 & 0.83 & 0.80 & \textbf{0.90} & 0.79 \\
    
    \midrule
    \multirow{3}{*}{Hypertension} 
    & RMSE & 6.70 & 12.34 & 9.88 & 8.51 & 7.12 & 11.64 \\
    & MAE  & 4.55 & 7.86 & 6.81 & 5.69 & 5.17 & 7.31 \\
    & $R^2$ & 0.83 & 0.81 & 0.87 & 0.84 & \textbf{0.93} & 0.73 \\

    \midrule
    \multirow{3}{*}{Asthma} 
    & RMSE & 1.14 & 1.55 & 1.41 & 1.49 & 1.25 & 1.62 \\
    & MAE  & 0.87 & 1.03 & 0.93 & 1.13 & 0.89 & 1.11 \\
    & $R^2$ & 0.76 & 0.82 & 0.84 & 0.76 & \textbf{0.88} & 0.74 \\

    \midrule
    \multirow{3}{*}{Depression} 
    & RMSE & 1.98 & 2.53 & 2.54 & 2.63 & 2.82 & 2.68 \\
    & MAE  & 1.37 & 1.89 & 1.65 & 1.90 & 1.74 & 1.84 \\
    & $R^2$ & 0.89 & \textbf{0.90} & 0.88 & 0.87 & 0.88 & 0.77 \\

    \midrule
    \multirow{3}{*}{Anxiety} 
    & RMSE & 3.81 & 3.70 & 4.27 & 3.83 & 3.49 & 4.49 \\
    & MAE  & 2.39 & 2.70 & 2.87 & 2.83 & 2.32 & 2.82 \\
    & $R^2$ & 0.81 & 0.89 & 0.84 & 0.87 & \textbf{0.90} & 0.75 \\

    \midrule
    \multirow{3}{*}{Opioids} 
    & RMSE & 1.44 & 2.28 & 1.62 & 1.99 & 1.98 & 1.63 \\
    & MAE  & 0.94 & 1.50 & 1.09 & 1.37 & 1.45 & 1.13 \\
    & $R^2$ & 0.83 & 0.84 & 0.85 & 0.77 & \textbf{0.89} & 0.76 \\

    \bottomrule
  \end{tabular}
\end{table}

Compared with the 10-fold cross-validation, UST-GNN experienced a slight performance decline under LOOSCV for most outcomes, with $R^2$ scores generally reduced by less than 0.1. A more pronounced drop was observed for diabetes in Camden and Ealing, which likely reflects greater generalisation difficulty under borough-level hold-out. %rather than a single common cause.
Plausible explanations include distribution shift between held-out and training boroughs, variation in feature--outcome relationships across boroughs, and higher intra-borough heterogeneity. Similarly, asthma prediction in Camden, Southwark, and Ealing showed reduced $R^2$, suggesting that neighbourhood-level patterns in these boroughs were more heterogeneous or less well represented by the current graph attributes.

Despite these expected reductions, UST-GNN still maintained an average $R^2$ above 0.80 across all outcomes and boroughs. Notably, prediction in Croydon consistently achieved the highest scores, indicating that certain boroughs may have more regular socio-environmental patterns that the model can capture more effectively under spatial hold-out.

Overall, even under this more stringent validation regime, where entire unseen Borough subgraphs were held out, UST-GNN continued to outperform traditional baselines (SLM, GWR, LightGBM) evaluated under the less challenging 10-fold setting (Table~\ref{tab:loocv_results}). This suggests that UST-GNN remains robust under heterogeneous spatial configurations, although borough-specific generalisation difficulty is clearly non-uniform.

\subsection{Interpretability Analysis}\label{sec:res_XAI}
\begin{figure}[!htbp]
    \centering
    \includegraphics[width=1\textwidth]{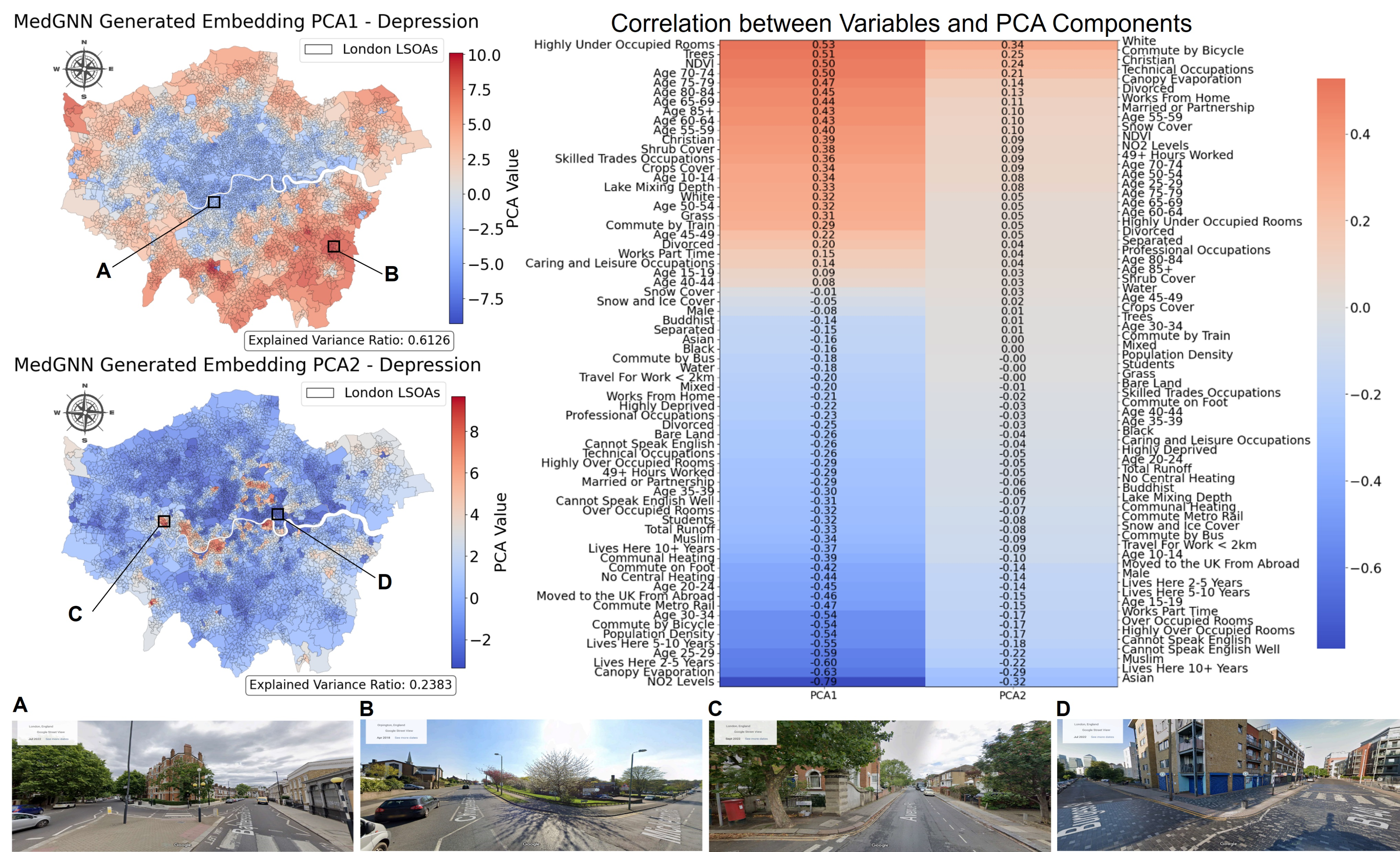}
    \caption{Principal Component Interpretation of learned embeddings and their explanatory links to antidepressant spatial patterns. (Top left) Spatial distributions of PCA1 and PCA2 across London LSOAs. (Top right) Correlation heatmap between embedding components and original socio-environmental features. (Bottom A–D) Representative street-view samples from annotated neighbourhoods.}
    \label{fig:pca_map}
\end{figure}

Aligned with our third contribution, we applied our proposed Principal Component Interpretation framework to the learned node embeddings from the depression prescription task. As shown in Figure~\ref{fig:pca_map}, the first two components (PCA1 and PCA2) explained 61.26\% and 23.83\% of the total variance, respectively, capturing the dominant spatial and socio-environmental structures embedded by UST-GNN.

The spatial maps of PCA1 and PCA2 revealed distinct but complementary patterns. PCA1 expressed a pronounced urban–rural gradient, with low values in central boroughs (e.g., Region A) and higher values in outer boroughs (e.g., Region B), aligning with known socio-demographic centrality. PCA2, in contrast, reflected intra-urban heterogeneity (e.g., Regions C and D), capturing finer distinctions among neighbourhoods that traditional spatial regression models often fail to represent.

The correlation heatmap between embedding components and MedSAT features (Figure~\ref{fig:pca_map}) indicated that PCA1 was negatively correlated with population density (\(-0.54\)), air pollution (\(-0.79\)), and income levels (\(-0.23\)), and positively correlated with housing deprivation (\(0.53\)). Together, these associations suggest that PCA1 captures a latent gradient related to urban inequality and environmental stress. PCA2 was associated with cultural and demographic variation, showing positive correlations with the proportion of White residents (\(0.34\)) and Christian beliefs (\(0.24\)), and negative correlations with Asian residents (\(-0.32\)) and Muslim beliefs (\(-0.22\)). It further captured built-environment attributes, such as housing quality and commuting modes, indicative of contextual differences in access and exposure.

Compared with established interpretability techniques such as Permutation-based attribution, Integrated Gradients, and GNNExplainer (see~\ref{app:explainability}), our PCA-embedding approach yielded more consistent and spatially continuous interpretations. It effectively summarised global structure–function relationships, offering a stable, transferable understanding of how the model internalises spatial context—thereby advancing interpretability in spatial graph learning beyond purely local attribution methods. Together, these findings demonstrate that UST-GNN embeddings encode interpretable, spatially coherent dimensions linking socio-environmental structure to health outcomes. 
\section{Discussion}
\subsection{UST-GNN: Model Insights and General Implications}
The proposed UST-GNN framework consistently outperformed all baseline models across six distinct prescription outcomes, demonstrating its ability to capture spatial dependencies, topological structures, and feature heterogeneity in complex urban systems. By combining expressive graph-based aggregation with complementary positional and location encodings, UST-GNN generated neighbourhood representations that were both more accurate and more spatially coherent than those produced by traditional spatial regression or tree-based machine learning approaches. Consistent with the view that “if a model approximates the data-generating process well, its interpretation should yield better insights into that process”~\citep{molnar2020pitfalls}, our results showed that UST-GNN not only improved predictive accuracy but also uncovered interpretable latent structures that aligned with known spatial inequalities.

A key contribution of this study lies in the design of a transferable feature engineering pipeline for high-dimensional urban datasets. Existing studies often mitigate multicollinearity through simple VIF filtering or single-metric ranking~\citep{cheng2022variable}, yet such approaches overlook the compounded dependencies among socio-demographic and environmental variables~\citep{trushna2021effects,wu2023measuring}. Our comprehensive multi-dimensional feature selection framework (~\ref{app:feature_selection}) integrates four complementary evaluation metrics, mutual information, F-statistics, random forest importance, and Lasso regularization, within an adaptive cross-validation procedure and correlation-based redundancy control ($|r|>0.8$). 
The approach balances theoretical relevance with statistical robustness by mandatorily retaining core control variables (e.g., demographics, deprivation, environmental exposure) while optimizing the remaining feature set in a data-driven manner. The resulting subset achieves near-saturated explanatory capacity with substantially reduced dimensionality, providing a stable, interpretable, and generalizable representation of urban socio-environmental systems. This integrated pipeline is broadly applicable to other spatial prediction and health geography tasks, supporting reproducible, domain-grounded feature learning across diverse urban contexts.

Beyond performance gains, the systematic spatial component ablation (covering graph architecture, spatial adjacency definitions, neighbourhood depth, and positional encoding) reinforced that spatially explicit GNN effectiveness depends as much on how spatial relationships are encoded as on the neural architecture itself. While automated methods such as Neural Architecture Search (NAS)~\citep{zhou2022auto} may further refine configurations, their computational demands often limit applicability in high-resolution urban datasets. Our sequential ablation procedure offers a more tractable yet still rigorous pathway for spatial model optimisation.

More broadly, our findings highlight a methodological point for GeoAI-based urban analytics: spatial effects, network topology, leakage-aware evaluation, and interpretability are often treated as separate modelling concerns, even though they jointly shape the reliability and usefulness of urban prediction systems. In policy-facing settings such as urban health, predictive performance alone is rarely sufficient; models must also generate explanations that are spatially meaningful and interpretable at a level relevant to intervention and planning. By integrating these elements within a single spatially explicit unified framework, UST-GNN moves beyond prediction toward insight generation, showing that graph-based urban models can support both robust forecasting and policy-relevant interpretation. In this sense, the framework also provides a useful foundation for future urban digital twin workflows, where heterogeneous urban data, scenario testing, monitoring, and transparent analytical outputs must be combined to support evidence-based decision-making.

Finally, although evaluated here in the context of health outcomes, UST-GNN is outcome-agnostic and readily transferable to other neighbourhood-level prediction tasks, ranging from housing prices and energy demand to accessibility, environmental risk, and socio-economic deprivation. This flexibility underscores the potential of the framework as a reusable GeoAI pipeline for diverse urban analytics applications and for broader evidence-based planning in sustainable and equitable cities.

\subsection{Interpretation through PCA-embedding method}

To interpret the deep representation learned by UST-GNN, we applied PCA to the final-layer node embeddings (Figure~\ref{fig:pca_map}). The first two components (PCA1 and PCA2) explained 61.3\% and 23.8\% of the total variance, respectively, capturing the dominant spatial and structural gradients encoded by the network. PCA1 exhibited a pronounced centre–periphery gradient consistent with established urban form and demographic structures in Greater London~\citep{hatch2011identifying,aiello2019large}, with high scores linked to denser, younger, and less green areas, and low scores characterising suburban and greener neighbourhoods. PCA2 captured finer-scale socio-cultural variations, identifying clusters of inner-London communities with elevated antidepressant prescription rates and strong associations with ethnicity and religion.

Regression of the observed antidepressant prescription rates on PCA1 and PCA2 yielded \( R^2 > 0.80 \), indicating that these components collectively explained the majority of spatial variation captured by UST-GNN. Together, PCA1 and PCA2 revealed complementary concepts: the first represented macro-level structural–environmental gradients, while the second seemed to exhibit localised socio-cultural contrasts underlying heterogeneous mental health outcomes.

Unlike conventional attribution-based explainers such as SHAP, Integrated Gradients, or GNNExplainer (see ~\ref{app:explainability}), our Principal Component Interpretation analysis does not attempt to assign importance to individual input features, nor does it seek  to disentangled neurons like mechanistic interpretability. Instead, it focuses on identifying and interpreting the most informative and intrinsic component of the model's learned representation. This inductive, correlation-oriented approach trades off granular instance-level precision for global interpretability and computational efficiency, an advantage in high-dimensional, spatially dependent systems where strict feature-level causality is often ill-posed. While acknowledging limitations with analysing these entangled, concept-level components, this method reveals spatial gradients linking \textit{raw features, learned embeddings, and target outcomes}, a level of structural insight that most local attribution methods cannot provide.

In this sense, the Principal Component Interpretation framework serves as a scalable and domain-grounded interpretability strategy: it transforms complex GNN representations into policy-relevant axes of socio-environmental variation, facilitating cross-model comparison and rapid exploration of structure–function relationships in urban systems. This approach, though deliberately simplified, provides a transferable and analytically transparent pathway to understanding how deep spatial models generalise across diverse geographic contexts.

\subsubsection{Alignment with Existing Literature}
%%%%%%%%%%%%%%%%%%%%%%%%%%%%%%%%%%%%%%%%%%%

Our results affirm prior findings regarding age, ethnicity, and housing conditions in relation to mental health outcomes. Antidepressant prescription rates were highest among elderly populations~\citep{cahoon2012depression} and adolescents aged 10–19~\citep{lalji2021analysis,armitage2021antidepressants}, and lower among adults aged 20–39. This younger adult group tends to exhibit health-protective behaviors such as active commuting~\citep{berrie2024does,soini2024physical}, professional employment~\citep{remes2021biological}, and remote work—an arrangement shown to buffer mental distress, especially during the COVID-19 pandemic~\citep{bonacini2021working}.

Notably, PCA1 showed strong correlation with under-occupied housing (correlation coefficient: 0.53), a potential proxy for social isolation and loneliness~\citep{erzen2018effect}. Both PCA1 and PCA2 also exhibited positive associations with the proportion of White residents, consistent with prior evidence suggesting higher antidepressant usage in this demographic~\citep{sclar2008ethnicity}.

Conversely, lower prescription rates were observed in areas with high proportions of Muslim residents, recent immigrants, and individuals with limited English proficiency. These trends may reflect not only cultural stigma or differing health beliefs~\citep{loewenthal2002women,assari2017social}, but also structural barriers to care. For instance, language limitations have been identified as a barrier to receiving appropriate depression treatment~\citep{lesser2008depression}. This suggests that observed disparities may arise from inequities in access rather than differences in need. Figures~\ref{fig:pca_map} A–D illustrate representative neighbourhood typologies from these areas.

\subsubsection{Adding to Ongoing Debates}
%%%%%%%%%%%%%%%%%%%%%%%%%%%%%%%%%%%%%%%%%%%

Our findings contribute to several ongoing debates within the urban health literature by challenging commonly accepted associations and highlighting context-specific dynamics.

First, we observe a positive correlation between green land cover (e.g., crops, shrubland) and antidepressant prescriptions. While most research links exposure to green spaces with lower depression incidence~\citep{browning2019tree}, some recent studies have reported similar counterintuitive trends ~\citep{astell2022urban,hyam2020greenness}. These discrepancies may reflect heterogeneity in green space accessibility, usage patterns, or the socio-demographic composition of green areas. For example, peripheral neighborhoods rich in vegetation may coincide with older or more isolated populations, partially explaining the observed increase in prescriptions.

Second, working more than 49 hours per week (exceeding the UK Working Time Directive\footnote{\href{https://www.gov.uk/maximum-weekly-working-hours}{UK Government guidance on maximum weekly working hours}}) is negatively correlated with antidepressant prescriptions (\(r = -0.29\) with PCA1). This pattern differs from prior literature associating long working hours with elevated depression risk~\citep{melchior2007work}. One possible interpretation is that this variable may proxy for correlated factors such as stable employment and income, both of which are often associated with better mental health outcomes. This aligns with newer perspectives suggesting that certain types of work-related stress may be positively associated with purpose and psychological resilience~\citep{vscepanovic2023quantifying}.

Third, we find a negative correlation between NO\textsubscript{2} levels and antidepressant use, a result that diverges from the prevailing consensus linking air pollution to adverse mental health outcomes~\citep{lim2012air}. Similar trends were observed in other geospatial studies~\citep{fleury2021geospatial} and within the MedSAT dataset itself~\citep{scepanovic2024medsat,maitra2025location}. A possible explanation lies in urban policy interventions: London’s Ultra Low Emission Zone (ULEZ), introduced in stages since 2019\footnote{\href{https://tfl.gov.uk/modes/driving/ultra-low-emission-zone}{Transport for London Ultra Low Emission Zone policy}}, may have spatially decoupled pollution levels from mental health burdens by redistributing high-emission zones away from vulnerable populations.

\subsubsection{Insights Warranting Future Research}
%%%%%%%%%%%%%%%%%%%%%%%%%%%%%%%%%%%%%%%%%%%

UST-GNN’s pipeline, coupled with the comprehensive MedSAT dataset, enables a comprehensive exploration of urban-environment-health interrelations beyond the scope of conventional public health analyses. This modelling approach surfaced several previously underexplored associations that merit further investigation.

Among them, we find a positive correlation between antidepressant prescriptions and both train commuting rates and employment in caring and leisure occupations. While public transport and caregiving roles are typically viewed as health-supportive due to increased accessibility and social engagement, their observed links with higher prescription rates may reflect context-dependent stressors such as emotional labor, shift work, or pandemic-related occupational burnout. These findings suggest that presumed protective urban or occupational features may carry hidden mental health burdens depending on exposure intensity and socio-environmental context.

Furthermore, environmental hydrological features, specifically-total surface runoff and canopy evaporation—show consistent negative correlations with antidepressant prescriptions. Total runoff, which quantifies the volume of precipitation-induced surface water flow, may serve as a proxy for effective stormwater infrastructure and urban green design, often characteristic of walkable, eco-resilient neighborhoods. Similarly, higher canopy evaporation, primarily driven by extensive tree cover, contributes to urban cooling and microclimatic regulation, mitigating thermal stress and promoting psychological comfort. These environmental buffers may form indirect yet impactful pathways for mental well-being in dense urban settings.

Together, these associations highlight the ability of graph-based spatial machine learning to reveal hidden structural and environmental drivers of urban health. Advancing this work requires transdisciplinary collaboration between urban science, causal inference, and public health to connect model-derived hypotheses with individual-level data, qualitative interviews and targeted interventions. Integrating such insights into policy can help shape more equitable and responsive urban health environments.

\section{Conclusion}
This study demonstrates the effectiveness of a unified spatio-topological graph neural network (UST-GNN) for neighbourhood-level health outcome prediction from rich socio-demographic and environmental datasets. In a London case study on health prescriptions, UST-GNN consistently outperformed statistical, machine learning and geographically enriched baselines by jointly modelling nonlinear spatial dependencies, network topology, with heterogenous features in strict spatial cross validation settings. Principal component interpretation analysis further demonstrated that the model not only achieved high predictive accuracy but also produced interpretable and meaningful links to real-world urban patterns that validated established drivers, contributed to ongoing debates, and revealed novel associations warranting further causal investigation.

While demonstrated here in an urban health context, UST-GNN is inherently outcome-agnostic and can be readily applied to a wide range of spatial prediction tasks, such as modelling energy demand, housing prices, or environmental risk indices, offering a reproduceable GeoAI pipeline for urban analytics, spatial epidemiology, and equitable city planning. Beyond these applications, the proposed framework can serve as an analytical pipeline within emerging \textit{urban digital twin} systems, that continuously integrate real-time data for monitoring, simulation, and policy experimentation~\citep{malleson2024digital}. Embedding UST-GNN into digital-twin workflows (eg. agent-based-models) would enable data-driven equitable urban planning to foster healthier and more sustainable cities.

%% ================== BIBLIOGRAPHY ===================
% \bibliographystyle{elsarticle-num-names}
 \bibliographystyle{elsarticle-harv}
\bibliography{bibliography}

@article{scepanovic2024medsat,
  title={MedSat: A Public Health Dataset for England Featuring Medical Prescriptions and Satellite Imagery},
  author={\v{S}\'{c}epanovi\'{c}, Sanja and Obadic, Ivica and Joglekar, Sagar and Giustarini, Laura and Nattero, Cristiano and Quercia, Daniele and Zhu, Xiaoxiang},
  journal={Advances in Neural Information Processing Systems},
  volume={36},
  year={2024}
}

@article{cole2021breaking,
  title={Breaking down and building up: gentrification, its drivers, and urban health inequality},
  author={Cole, H. V. and Mehdipanah, R. and Gullón, P. and Triguero-Mas, M.},
  journal={Current Environmental Health Reports},
  volume={8},
  pages={157--166},
  year={2021},
  publisher={Springer}
}

@article{obradovich2018empirical,
  title={Empirical evidence of mental health risks posed by climate change},
  author={Obradovich, N. and Migliorini, R. and Paulus, M. P. and Rahwan, I.},
  journal={Proceedings of the National Academy of Sciences},
  volume={115},
  number={43},
  pages={10953--10958},
  year={2018},
  publisher={National Academy of Sciences}
}

@article{fiscella2004health,
  title={Health disparities based on socioeconomic inequities: implications for urban health care},
  author={Fiscella, K. and Williams, D. R.},
  journal={Academic Medicine},
  volume={79},
  number={12},
  pages={1139--1147},
  year={2004},
  publisher={Lippincott Williams \& Wilkins}
}

@article{maitra2025location,
  title={How Your Location Relates to Health: Variable Importance and Interpretable Machine Learning for Environmental and Sociodemographic Data},
  author={Maitra, I. and Lin, R. and Chen, E. and Donnelly, J. and {\v{S}}ćepanovi{\'c}, S. and Rudin, C.},
  journal={arXiv preprint arXiv:2501.02111},
  year={2025}
}

@article{li2023survey,
  title={A survey of graph neural network based recommendation in social networks},
  author={Li, X. and Sun, L. and Ling, M. and Peng, Y.},
  journal={Neurocomputing},
  volume={549},
  pages={126441},
  year={2023},
  publisher={Elsevier}
}

@inproceedings{sun2022gnn,
  title={Does GNN pretraining help molecular representation?},
  author={Sun, R. and Dai, H. and Yu, A. W.},
  booktitle={Advances in Neural Information Processing Systems},
  volume={35},
  pages={12096--12109},
  year={2022}
}

@article{wu2022graph,
  title={Graph neural networks in recommender systems: a survey},
  author={Wu, S. and Sun, F. and Zhang, W. and Xie, X. and Cui, B.},
  journal={ACM Computing Surveys},
  volume={55},
  number={5},
  pages={1--37},
  year={2022},
  publisher={ACM}
}

@article{liu2024explainable,
  title={Explainable spatially explicit geospatial artificial intelligence in urban analytics},
  author={Liu, P. and Zhang, Y. and Biljecki, F.},
  journal={Environment and Planning B: Urban Analytics and City Science},
  volume={51},
  number={5},
  pages={1104--1123},
  year={2024},
  publisher={SAGE Publications}
}

@article{zhang2023knowledge,
  title={Knowledge and topology: A two layer spatially dependent graph neural networks to identify urban functions with time-series street view image},
  author={Zhang, Y. and Liu, P. and Biljecki, F.},
  journal={ISPRS Journal of Photogrammetry and Remote Sensing},
  volume={198},
  pages={153--168},
  year={2023},
  publisher={Elsevier}
}

@article{liu2024association,
  title={The association between urban density and multiple health risks based on interpretable machine learning: A study of American urban communities},
  author={Liu, Z. and Liu, C.},
  journal={Cities},
  volume={153},
  pages={105170},
  year={2024},
  publisher={Elsevier}
}

@article{kjellstrom2016impact,
  title={Impact of climate conditions on occupational health and related economic losses: a new feature of global and urban health in the context of climate change},
  author={Kjellstrom, T.},
  journal={Asia Pacific Journal of Public Health},
  volume={28},
  number={2\_suppl},
  pages={28S--37S},
  year={2016},
  publisher={SAGE Publications}
}

@article{kedzie2018content,
  title={Content selection in deep learning models of summarization},
  author={Kedzie, C. and McKeown, K. and Daume III, H.},
  journal={arXiv preprint arXiv:1810.12343},
  year={2018}
}

@article{adnan2022utilizing,
  title={Utilizing grid search cross-validation with adaptive boosting for augmenting performance of machine learning models},
  author={Adnan, M. and Alarood, A. A. S. and Uddin, M. I. and ur Rehman, I.},
  journal={PeerJ Computer Science},
  volume={8},
  pages={e803},
  year={2022},
  publisher={PeerJ}
}

@article{greenacre2022principal,
  title={Principal component analysis},
  author={Greenacre, M. and Groenen, P. J. and Hastie, T. and d’Enza, A. I. and Markos, A. and Tuzhilina, E.},
  journal={Nature Reviews Methods Primers},
  volume={2},
  number={1},
  pages={100},
  year={2022},
  publisher={Nature Publishing Group}
}

@book{cohen2013applied,
  title={Applied multiple regression/correlation analysis for the behavioral sciences},
  author={Cohen, J. and Cohen, P. and West, S. G. and Aiken, L. S.},
  year={2013},
  publisher={Routledge}
}

@article{zhang2019graph,
  title={Graph convolutional networks: a comprehensive review},
  author={Zhang, S. and Tong, H. and Xu, J. and Maciejewski, R.},
  journal={Computational Social Networks},
  volume={6},
  number={1},
  pages={1--23},
  year={2019},
  publisher={Springer}
}

@article{bhatti2023deep,
  title={Deep learning with graph convolutional networks: An overview and latest applications in computational intelligence},
  author={Bhatti, U. A. and Tang, H. and Wu, G. and Marjan, S. and Hussain, A.},
  journal={International Journal of Intelligent Systems},
  volume={2023},
  number={1},
  pages={8342104},
  year={2023},
  publisher={Wiley}
}

@article{kipf2016semi,
  title={Semi-supervised classification with graph convolutional networks},
  author={Kipf, T. N. and Welling, M.},
  journal={arXiv preprint arXiv:1609.02907},
  year={2016}
}

@misc{ons2025,
  author = {{Office for National Statistics (ONS)}},
  title = {National Statistics},
  year = {2025},
  howpublished = {\url{https://www.ons.gov.uk}},
  note = {Accessed: 2025-06-12}
}

@misc{nhs2025,
  author = {{National Health Service (NHS)}},
  title = {NHS Health Data},
  year = {2025},
  howpublished = {\url{https://www.nhs.uk}},
  note = {Accessed: 2025-06-12}
}

@misc{gee2025,
  author = {{Google Earth Engine (GEE)}},
  title = {Google Earth Engine: A Cloud-based Platform for Earth Science Data Analysis},
  year = {2025},
  howpublished = {\url{https://earthengine.google.com}},
  note = {Accessed: 2025-06-12}
}

@inproceedings{chen2020iterative,
  title={Iterative deep graph learning for graph neural networks: Better and robust node embeddings},
  author={Chen, Y. and Wu, L. and Zaki, M.},
  booktitle={Advances in Neural Information Processing Systems},
  volume={33},
  pages={19314--19326},
  year={2020}
}

@inproceedings{feng2022powerful,
  title={How powerful are k-hop message passing graph neural networks},
  author={Feng, J. and Chen, Y. and Li, F. and Sarkar, A. and Zhang, M.},
  booktitle={Advances in Neural Information Processing Systems},
  volume={35},
  pages={4776--4790},
  year={2022}
}

@article{klemmer2023satclip,
  title={Satclip: Global, general-purpose location embeddings with satellite imagery},
  author={Klemmer, K. and Rolf, E. and Robinson, C. and Mackey, L. and Ru{\ss}wurm, M.},
  journal={arXiv preprint arXiv:2311.17179},
  year={2023}
}

@article{schaub2020random,
  title={Random walks on simplicial complexes and the normalized Hodge 1-Laplacian},
  author={Schaub, M. T. and Benson, A. R. and Horn, P. and Lippner, G. and Jadbabaie, A.},
  journal={SIAM Review},
  volume={62},
  number={2},
  pages={353--391},
  year={2020},
  publisher={SIAM}
}

@article{fushiki2011estimation,
  title={Estimation of prediction error by using K-fold cross-validation},
  author={Fushiki, T.},
  journal={Statistics and Computing},
  volume={21},
  pages={137--146},
  year={2011},
  publisher={Springer}
}

@article{wong2015performance,
  title={Performance evaluation of classification algorithms by k-fold and leave-one-out cross validation},
  author={Wong, T. T.},
  journal={Pattern Recognition},
  volume={48},
  number={9},
  pages={2839--2846},
  year={2015},
  publisher={Elsevier}
}

@article{kihal2013green,
  title={Green space, social inequalities and neonatal mortality in France},
  author={Kihal-Talantikite, Wahida and Padilla, Clara M and Lalloué, Benoit and Gelormini, Marcella and Zmirou-Navier, Denis and Deguen, Severine},
  journal={BMC Pregnancy and Childbirth},
  volume={13},
  pages={1--9},
  year={2013},
  publisher={BioMed Central}
}

@article{zhou2023combined,
  title={Combined effects of heatwaves and air pollution, green space and blue space on the incidence of hypertension: a national cohort study},
  author={Zhou, Wei and Wang, Qi and Li, Rong and Kadier, Abliz and Wang, Wenjing and Zhou, Feifei and Ling, Li},
  journal={Science of The Total Environment},
  volume={867},
  pages={161560},
  year={2023},
  publisher={Elsevier}
}

@article{white2021associations,
  title={Associations between green/blue spaces and mental health across 18 countries},
  author={White, Mathew P. and Elliott, Lewis R. and Grellier, James and Economou, Tonia and Bell, Sarah and Bratman, Gregory N. and Fleming, Lora E.},
  journal={Scientific Reports},
  volume={11},
  number={1},
  pages={8903},
  year={2021},
  publisher={Nature Publishing Group}
}

@article{mazumdar2021green,
  title={Which green space metric best predicts a lowered odds of type 2 diabetes?},
  author={Mazumdar, Sharmistha and Chong, Sylvia and Astell-Burt, Thomas and Feng, Xiaoqi and Morgan, Geoff and Jalaludin, Bin},
  journal={International Journal of Environmental Research and Public Health},
  volume={18},
  number={8},
  pages={4088},
  year={2021},
  publisher={MDPI}
}

@article{van2020socioeconomic,
  title={Socioeconomic marginalization and opioid-related overdose: a systematic review},
  author={Van Draanen, Jenna and Tsang, Calvin and Mitra, Saakshi and Karamouzian, Mohammad and Richardson, Lindsey},
  journal={Drug and alcohol dependence},
  volume={214},
  pages={108127},
  year={2020},
  publisher={Elsevier}
}

@article{ohanyan2022associations,
  title={Associations between the urban exposome and type 2 diabetes: Results from penalised regression by least absolute shrinkage and selection operator and random forest models},
  author={Ohanyan, H. and Portengen, L. and Kaplani, O. and Huss, A. and Hoek, G. and Beulens, J. W. and Vermeulen, R.},
  journal={Environment International},
  volume={170},
  pages={107592},
  year={2022},
  publisher={Elsevier}
}

@article{rothenberg2014flexible,
  title={A flexible urban health index for small area disparities},
  author={Rothenberg, R. and Weaver, S. R. and Dai, D. and Stauber, C. and Prasad, A. and Kano, M.},
  journal={Journal of Urban Health},
  volume={91},
  number={5},
  pages={823--835},
  year={2014},
  publisher={Springer}
}

@article{shane2000urban,
  title={Urban environmental sustainability metrics: a provisional set},
  author={Shane, A. M. and Graedel, T. E.},
  journal={Journal of Environmental Planning and Management},
  volume={43},
  number={5},
  pages={643--663},
  year={2000},
  publisher={Taylor \& Francis}
}

@inproceedings{song2004comparison,
  title={Comparison of machine learning techniques with classical statistical models in predicting health outcomes},
  author={Song, X. and Mitnitski, A. and Cox, J. and Rockwood, K.},
  booktitle={MEDINFO 2004},
  pages={736--740},
  year={2004},
  publisher={IOS Press}
}

@article{pineo2018urban,
  title={Urban health indicator tools of the physical environment: a systematic review},
  author={Pineo, H. and Glonti, K. and Rutter, H. and Zimmermann, N. and Wilkinson, P. and Davies, M.},
  journal={Journal of Urban Health},
  volume={95},
  pages={613--646},
  year={2018},
  publisher={Springer}
}

@article{jordan2015machine,
  title={Machine learning: Trends, perspectives, and prospects},
  author={Jordan, M. I. and Mitchell, T. M.},
  journal={Science},
  volume={349},
  number={6245},
  pages={255--260},
  year={2015},
  publisher={AAAS}
}

@article{bhakta2016prediction,
  title={Prediction of depression among senior citizens using machine learning classifiers},
  author={Bhakta, I. and Sau, A.},
  journal={International Journal of Computer Applications},
  volume={144},
  number={7},
  pages={11--16},
  year={2016},
  publisher={Foundation of Computer Science}
}

@article{su2021use,
  title={Use of machine learning approach to predict depression in the elderly in China: a longitudinal study},
  author={Su, D. and Zhang, X. and He, K. and Chen, Y.},
  journal={Journal of Affective Disorders},
  volume={282},
  pages={289--298},
  year={2021},
  publisher={Elsevier}
}

@article{thotad2023diabetes,
  title={Diabetes disease detection and classification on Indian demographic and health survey data using machine learning methods},
  author={Thotad, P. N. and Bharamagoudar, G. R. and Anami, B. S.},
  journal={Diabetes \& Metabolic Syndrome: Clinical Research \& Reviews},
  volume={17},
  number={1},
  pages={102690},
  year={2023},
  publisher={Elsevier}
}

@article{arcaya2016research,
  title={Research on neighborhood effects on health in the United States: a systematic review of study characteristics},
  author={Arcaya, M. C. and Tucker-Seeley, R. D. and Kim, R. and Schnake-Mahl, A. and So, M. and Subramanian, S. V.},
  journal={Social Science \& Medicine},
  volume={168},
  pages={16--29},
  year={2016},
  publisher={Elsevier}
}

@article{sarkar2017urban,
  title={Urban environments and human health: Current trends and future directions},
  author={Sarkar, C. and Webster, C.},
  journal={Current Opinion in Environmental Sustainability},
  volume={25},
  pages={33--44},
  year={2017},
  publisher={Elsevier}
}

@article{lalloue2013statistical,
  title={A statistical procedure to create a neighborhood socioeconomic index for health inequalities analysis},
  author={Lalloué, B. and Monnez, J. M. and Padilla, C. and Kihal, W. and Le Meur, N. and Zmirou-Navier, D. and Deguen, S.},
  journal={International Journal for Equity in Health},
  volume={12},
  pages={1--11},
  year={2013},
  publisher={BioMed Central}
}

@article{scheider2023pragmatic,
  title={Pragmatic GeoAI: Geographic information as externalized practice},
  author={Scheider, S. and Richter, K. F.},
  journal={KI-Künstliche Intelligenz},
  volume={37},
  number={1},
  pages={17--31},
  year={2023},
  publisher={Springer}
}

@article{vopham2018emerging,
  title={Emerging trends in geospatial artificial intelligence (GeoAI): potential applications for environmental epidemiology},
  author={VoPham, T. and Hart, J. E. and Laden, F. and Chiang, Y. Y.},
  journal={Environmental Health},
  volume={17},
  pages={1--6},
  year={2018},
  publisher={BioMed Central}
}

@article{li2021prediction,
  title={Prediction of human activity intensity using the interactions in physical and social spaces through graph convolutional networks},
  author={Li, M. and Gao, S. and Lu, F. and Liu, K. and Zhang, H. and Tu, W.},
  journal={International Journal of Geographical Information Science},
  volume={35},
  number={12},
  pages={2489--2516},
  year={2021},
  publisher={Taylor \& Francis}
}

@inproceedings{wu2021inductive,
  title={Inductive graph neural networks for spatiotemporal kriging},
  author={Wu, Y. and Zhuang, D. and Labbe, A. and Sun, L.},
  booktitle={Proceedings of the AAAI Conference on Artificial Intelligence},
  volume={35},
  number={5},
  pages={4478--4485},
  year={2021},
  publisher={AAAI Press}
}

@inproceedings{choi2020learning,
  title={Learning the graphical structure of electronic health records with graph convolutional transformer},
  author={Choi, E. and Xu, Z. and Li, Y. and Dusenberry, M. and Flores, G. and Xue, E. and Dai, A.},
  booktitle={Proceedings of the AAAI Conference on Artificial Intelligence},
  volume={34},
  number={01},
  pages={606--613},
  year={2020},
  publisher={AAAI Press}
}

@article{helbich2018toward,
  title={Toward dynamic urban environmental exposure assessments in mental health research},
  author={Helbich, M.},
  journal={Environmental Research},
  volume={161},
  pages={129--135},
  year={2018},
  publisher={Elsevier}
}

@article{tung2017spatial,
  title={Spatial context and health inequity: reconfiguring race, place, and poverty},
  author={Tung, E. L. and Cagney, K. A. and Peek, M. E. and Chin, M. H.},
  journal={Journal of Urban Health},
  volume={94},
  pages={757--763},
  year={2017},
  publisher={Springer}
}

@article{lu2021weighted,
  title={A weighted patient network-based framework for predicting chronic diseases using graph neural networks},
  author={Lu, H. and Uddin, S.},
  journal={Scientific Reports},
  volume={11},
  number={1},
  pages={22607},
  year={2021},
  publisher={Nature Publishing Group}
}

@article{fritz2022combining,
  title={Combining graph neural networks and spatio-temporal disease models to improve the prediction of weekly COVID-19 cases in Germany},
  author={Fritz, C. and Dorigatti, E. and Rügamer, D.},
  journal={Scientific Reports},
  volume={12},
  number={1},
  pages={3930},
  year={2022},
  publisher={Nature Publishing Group}
}

@article{mai2020multi,
  title={Multi-scale representation learning for spatial feature distributions using grid cells},
  author={Mai, G. and Janowicz, K. and Yan, B. and Zhu, R. and Cai, L. and Lao, N.},
  journal={arXiv preprint arXiv:2003.00824},
  year={2020}
}

@article{desabbata2023graph,
  title={A graph neural network framework for spatial geodemographic classification},
  author={De Sabbata, S. and Liu, P.},
  journal={International Journal of Geographical Information Science},
  volume={37},
  number={12},
  pages={2464--2486},
  year={2023},
  publisher={Taylor \& Francis}
}

@article{dwivedi2021graph,
  title={Graph neural networks with learnable structural and positional representations},
  author={Dwivedi, V. P. and Luu, A. T. and Laurent, T. and Bengio, Y. and Bresson, X.},
  journal={arXiv preprint arXiv:2110.07875},
  year={2021},
  url={https://arxiv.org/abs/2110.07875}
}

@inproceedings{yeh2023random,
  title={Random walk conformer: learning graph representation from long and short range},
  author={Yeh, P. K. and Chen, H. W. and Chen, M. S.},
  booktitle={Proceedings of the AAAI Conference on Artificial Intelligence},
  volume={37},
  number={9},
  pages={10936--10944},
  year={2023},
  month={June},
  publisher={AAAI Press}
}

@article{cheng2022variable,
  title={A variable selection method based on mutual information and variance inflation factor},
  author={Cheng, J. and Sun, J. and Yao, K. and Xu, M. and Cao, Y.},
  journal={Spectrochimica Acta Part A: Molecular and Biomolecular Spectroscopy},
  volume={268},
  pages={120652},
  year={2022},
  publisher={Elsevier}
}

@article{trushna2021effects,
  title={Effects of ambient air pollution on psychological stress and anxiety disorder: a systematic review and meta-analysis of epidemiological evidence},
  author={Trushna, T. and Dhiman, V. and Raj, D. and Tiwari, R. R.},
  journal={Reviews on Environmental Health},
  volume={36},
  number={4},
  pages={501--521},
  year={2021},
  publisher={De Gruyter}
}

@article{wu2023measuring,
  title={Measuring urban nighttime vitality and its relationship with urban spatial structure: A data-driven approach},
  author={Wu, C. and Zhao, M. and Ye, Y.},
  journal={Environment and Planning B: Urban Analytics and City Science},
  volume={50},
  number={1},
  pages={130--145},
  year={2023},
  publisher={SAGE Publications}
}

@article{meyes2019ablation,
  title={Ablation studies in artificial neural networks},
  author={Meyes, R. and Lu, M. and de Puiseau, C. W. and Meisen, T.},
  journal={arXiv preprint arXiv:1901.08644},
  year={2019},
  url={https://arxiv.org/abs/1901.08644}
}

@article{zhou2022auto,
  title={Auto-gnn: Neural architecture search of graph neural networks},
  author={Zhou, K. and Huang, X. and Song, Q. and Chen, R. and Hu, X.},
  journal={Frontiers in Big Data},
  volume={5},
  pages={1029307},
  year={2022},
  publisher={Frontiers}
}

@article{browning2019tree,
  title={Tree cover shows an inverse relationship with depressive symptoms in elderly residents living in US nursing homes},
  author={Browning, M. H. and Lee, K. and Wolf, K. L.},
  journal={Urban Forestry \& Urban Greening},
  volume={41},
  pages={23--32},
  year={2019},
  publisher={Elsevier}
}

@article{lim2012air,
  title={Air pollution and symptoms of depression in elderly adults},
  author={Lim, Y. H. and Kim, H. and Kim, J. H. and Bae, S. and Park, H. Y. and Hong, Y. C.},
  journal={Environmental Health Perspectives},
  volume={120},
  number={7},
  pages={1023--1028},
  year={2012},
  publisher={National Institute of Environmental Health Sciences}
}

@article{assari2017social,
  title={Social determinants of depression: The intersections of race, gender, and socioeconomic status},
  author={Assari, S.},
  journal={Brain Sciences},
  volume={7},
  number={12},
  pages={156},
  year={2017},
  publisher={MDPI}
}

@article{loewenthal2002women,
  title={Are women more religious than men? Gender differences in religious activity among different religious groups in the UK},
  author={Loewenthal, K. M. and MacLeod, A. K. and Cinnirella, M.},
  journal={Personality and Individual Differences},
  volume={32},
  number={1},
  pages={133--139},
  year={2002},
  publisher={Elsevier}
}

@article{vlahov2002urbanization,
  title={Urbanization, urbanicity, and health},
  author={Vlahov, D. and Galea, S.},
  journal={Journal of Urban Health},
  volume={79},
  pages={S1--S12},
  year={2002},
  publisher={Springer}
}

@article{bell2018changes,
  title={Changes in extreme events and the potential impacts on human health},
  author={Bell, J. E. and Brown, C. L. and Conlon, K. and Herring, S. and Kunkel, K. E. and Lawrimore, J. and Uejio, C.},
  journal={Journal of the Air \& Waste Management Association},
  volume={68},
  number={4},
  pages={265--287},
  year={2018},
  publisher={Taylor \& Francis}
}

@article{capolongo2020covid,
  title={COVID-19 and cities: From urban health strategies to the pandemic challenge. A decalogue of public health opportunities},
  author={Capolongo, S. and Rebecchi, A. and Buffoli, M. and Appolloni, L. and Signorelli, C. and Fara, G. M. and D'Alessandro, D.},
  journal={Acta Bio Medica: Atenei Parmensis},
  volume={91},
  number={2},
  pages={13},
  year={2020},
  publisher={Minerva Medica}
}

@inproceedings{ng1997preventing,
  author    = {Andrew Y. Ng},
  title     = {Preventing "Overfitting" of Cross-Validation Data},
  booktitle = {Proceedings of the 14th International Conference on Machine Learning (ICML 1997)},
  pages     = {245--253},
  year      = {1997},
  month     = {July},
  publisher = {Morgan Kaufmann},
}

@inproceedings{kearns1997algorithmic,
  author    = {Michael Kearns and Dana Ron},
  title     = {Algorithmic Stability and Sanity-Check Bounds for Leave-One-Out Cross-Validation},
  booktitle = {Proceedings of the Tenth Annual Conference on Computational Learning Theory (COLT 1997)},
  pages     = {152--162},
  year      = {1997},
  month     = {July},
  publisher = {ACM},
}

@article{hamilton2017inductive,
  title={Inductive representation learning on large graphs},
  author={Hamilton, Will and Ying, Zhitao and Leskovec, Jure},
  journal={Advances in neural information processing systems},
  volume={30},
  year={2017}
}

@article{bor2014child,
  title={Are child and adolescent mental health problems increasing in the 21st century? A systematic review},
  author={Bor, William and Dean, Angela J and Najman, Jacob and Hayatbakhsh, Reza},
  journal={Australian \& New Zealand journal of psychiatry},
  volume={48},
  number={7},
  pages={606--616},
  year={2014},
  publisher={Sage Publications Sage UK: London, England}
}

@article{harvey2017can,
  title={Can work make you mentally ill? A systematic meta-review of work-related risk factors for common mental health problems},
  author={Harvey, Samuel B and Modini, Matthew and Joyce, Sadhbh and Milligan-Saville, Josie S and Tan, Leona and Mykletun, Arnstein and Bryant, Richard A and Christensen, Helen and Mitchell, Philip B},
  journal={Occupational and environmental medicine},
  volume={74},
  number={4},
  pages={301--310},
  year={2017},
  publisher={BMJ Publishing Group Ltd}
}

@article{hossain2020epidemiology,
  title={Epidemiology of mental health problems in COVID-19: a review},
  author={Hossain, Md Mahbub and Tasnim, Samia and Sultana, Abida and Faizah, Farah and Mazumder, Hoimonty and Zou, Liye and McKyer, E Lisako J and Ahmed, Helal Uddin and Ma, Ping},
  journal={F1000Research},
  volume={9},
  year={2020},
  publisher={Faculty of 1000 Ltd}
}

@article{hynie2018social,
  title={The social determinants of refugee mental health in the post-migration context: A critical review},
  author={Hynie, Michaela},
  journal={The Canadian Journal of Psychiatry},
  volume={63},
  number={5},
  pages={297--303},
  year={2018},
  publisher={Sage Publications Sage CA: Los Angeles, CA}
}

@article{compton2015social,
  title={The social determinants of mental health},
  author={Compton, Michael T and Shim, Ruth S},
  journal={Focus},
  volume={13},
  number={4},
  pages={419--425},
  year={2015},
  publisher={Am Psychiatric Assoc}
}

@article{astell2022urban,
  title={Is urban green space associated with lower mental healthcare expenditure?},
  author={Astell-Burt, Thomas and Navakatikyan, Michael and Eckermann, Simon and Hackett, Maree and Feng, Xiaoqi},
  journal={Social Science \& Medicine},
  volume={292},
  pages={114503},
  year={2022},
  publisher={Elsevier}
}

@article{hyam2020greenness,
  title={Greenness, mortality and mental health prescription rates in urban Scotland-a population level, observational study},
  author={Hyam, Roger},
  journal={Research Ideas and Outcomes},
  volume={6},
  pages={e53542},
  year={2020},
  publisher={Pensoft Publishers}
}

@article{erzen2018effect,
  title={The effect of loneliness on depression: A meta-analysis},
  author={Erzen, Evren and {\c{C}}ikrikci, {\"O}zkan},
  journal={International Journal of Social Psychiatry},
  volume={64},
  number={5},
  pages={427--435},
  year={2018},
  publisher={Sage Publications Sage UK: London, England}
}

@article{lalji2021analysis,
  title={An analysis of antidepressant prescribing trends in England 2015--2019.},
  author={Lalji, Hasnain M and McGrogan, Anita and Bailey, Sarah J},
  journal={Journal of affective disorders reports},
  volume={6},
  pages={100205},
  year={2021},
  publisher={Elsevier}
}

@article{armitage2021antidepressants,
  title={Antidepressants, primary care, and adult mental health services in England during COVID-19},
  author={Armitage, Richard},
  journal={The Lancet Psychiatry},
  volume={8},
  number={2},
  pages={e3},
  year={2021},
  publisher={Elsevier}
}

@article{cahoon2012depression,
  title={Depression in older adults},
  author={Cahoon, Cynthia G},
  journal={AJN The American Journal of Nursing},
  volume={112},
  number={11},
  pages={22--30},
  year={2012},
  publisher={LWW}
}

@article{dos2019data,
  title={Data mining and machine learning techniques applied to public health problems: A bibliometric analysis from 2009 to 2018},
  author={dos Santos, Bruno Samways and Steiner, Maria Teresinha Arns and Fenerich, Amanda Trojan and Lima, Rafael Henrique Palma},
  journal={Computers \& Industrial Engineering},
  volume={138},
  pages={106120},
  year={2019},
  publisher={Elsevier}
}

@article{alegria2018social,
  title={Social determinants of mental health: where we are and where we need to go},
  author={Alegr{\'\i}a, Margarita and NeMoyer, Amanda and Falg{\`a}s Bagu{\'e}, Irene and Wang, Ye and Alvarez, Kiara},
  journal={Current psychiatry reports},
  volume={20},
  pages={1--13},
  year={2018},
  publisher={Springer}
}

@article{hayat2017statistical,
  title={Statistical methods used in the public health literature and implications for training of public health professionals},
  author={Hayat, Matthew J and Powell, Amanda and Johnson, Tessa and Cadwell, Betsy L},
  journal={PloS one},
  volume={12},
  number={6},
  pages={e0179032},
  year={2017},
  publisher={Public Library of Science San Francisco, CA USA}
}

@article{cousens2011alternatives,
  title={Alternatives to randomisation in the evaluation of public-health interventions: statistical analysis and causal inference},
  author={Cousens, Simon and Hargreaves, J and Bonell, Chris and Armstrong, B and Thomas, J and Kirkwood, BR and Hayes, R},
  journal={Journal of Epidemiology \& Community Health},
  volume={65},
  number={7},
  pages={576--581},
  year={2011},
  publisher={BMJ Publishing Group Ltd}
}

@article{ijeh2024predictive,
  title={Predictive modeling for disease outbreaks: a review of data sources and accuracy},
  author={Ijeh, Scholastica and Okolo, Chioma Anthonia and Arowoogun, Jeremiah Olawumi and Adeniyi, Adekunle Oyeyemi and Omotayo, Olufunke},
  journal={International Medical Science Research Journal},
  volume={4},
  number={4},
  pages={406--419},
  year={2024}
}

@article{remes2021biological,
  title={Biological, psychological, and social determinants of depression: a review of recent literature},
  author={Remes, Olivia and Mendes, Jo{\~a}o Francisco and Templeton, Peter},
  journal={Brain sciences},
  volume={11},
  number={12},
  pages={1633},
  year={2021},
  publisher={MDPI}
}

@article{hatch2011identifying,
  title={Identifying socio-demographic and socioeconomic determinants of health inequalities in a diverse London community: the South East London Community Health (SELCoH) study},
  author={Hatch, Stephani L and Frissa, Souci and Verdecchia, Maria and Stewart, Robert and Fear, Nicola T and Reichenberg, Abraham and Morgan, Craig and Kankulu, Bwalya and Clark, Jennifer and Gazard, Billy and others},
  journal={BMC public health},
  volume={11},
  pages={1--17},
  year={2011},
  publisher={Springer}
}

@article{aiello2019large,
  title={Large-scale and high-resolution analysis of food purchases and health outcomes},
  author={Aiello, Luca Maria and Schifanella, Rossano and Quercia, Daniele and Del Prete, Lucia},
  journal={EPJ Data Science},
  volume={8},
  number={1},
  pages={1--22},
  year={2019},
  publisher={SpringerOpen}
}

@article{fleury2021geospatial,
  title={Geospatial analysis of individual-based Parkinson's disease data supports a link with air pollution: A case-control study},
  author={Fleury, Vanessa and Himsl, Rebecca and Joost, St{\'e}phane and Nicastro, Nicolas and Bereau, Matthieu and Guessous, Idris and Burkhard, Pierre R},
  journal={Parkinsonism \& Related Disorders},
  volume={83},
  pages={41--48},
  year={2021},
  publisher={Elsevier}
}

@article{sclar2008ethnicity,
  title={Ethnicity/race and the diagnosis of depression and use of antidepressants by adults in the United States},
  author={Sclar, David A and Robison, Linda M and Skaer, Tracy L},
  journal={International clinical psychopharmacology},
  volume={23},
  number={2},
  pages={106--109},
  year={2008},
  publisher={LWW}
}

@article{berrie2024does,
  title={Does cycle commuting reduce the risk of mental ill-health? An instrumental variable analysis using distance to nearest cycle path},
  author={Berrie, Laurie and Feng, Zhiqiang and Rice, David and Clemens, Tom and Williamson, Lee and Dibben, Chris},
  journal={International journal of epidemiology},
  volume={53},
  number={1},
  pages={dyad153},
  year={2024},
  publisher={Oxford University Press}
}

@article{soini2024physical,
  title={Physical activity and specific symptoms of depression: A pooled analysis of six cohort studies},
  author={Soini, Eetu and Rosenstr{\"o}m, Tom and M{\"a}{\"a}tt{\"a}nen, Ilmari and Jokela, Markus},
  journal={Journal of Affective Disorders},
  volume={348},
  pages={44--53},
  year={2024},
  publisher={Elsevier}
}

@article{bonacini2021working,
  title={Working from home and income inequality: risks of a ‘new normal’with COVID-19},
  author={Bonacini, Luca and Gallo, Giovanni and Scicchitano, Sergio},
  journal={Journal of population economics},
  volume={34},
  number={1},
  pages={303--360},
  year={2021},
  publisher={Springer}
}

@article{lesser2008depression,
  title={Depression outcomes of Spanish-and English-speaking Hispanic outpatients in STAR* D},
  author={Lesser, Ira and Zisook, Sidney and Flores, Deborah and Sciolla, Andres and Wisniewski, Stephen and Cook, Ian and Epstein, Marcy and Rosales, Aurora and Gonzalez, Carlos and Trivedi, Madhukar and others},
  journal={Psychiatric Services},
  volume={59},
  number={11},
  pages={1273--1284},
  year={2008},
  publisher={Am Psychiatric Assoc}
}

@article{melchior2007work,
  title={Work stress precipitates depression and anxiety in young, working women and men},
  author={Melchior, Maria and Caspi, Avshalom and Milne, Barry J and Danese, Andrea and Poulton, Richie and Moffitt, Terrie E},
  journal={Psychological medicine},
  volume={37},
  number={8},
  pages={1119--1129},
  year={2007},
  publisher={Cambridge University Press}
}

@article{vscepanovic2023quantifying,
  title={Quantifying the impact of positive stress on companies from online employee reviews},
  author={{\v{S}}{\'c}epanovi{\'c}, Sanja and Constantinides, Marios and Quercia, Daniele and Kim, Seunghyun},
  journal={Scientific Reports},
  volume={13},
  number={1},
  pages={1603},
  year={2023},
  publisher={Nature Publishing Group UK London}
}

@article{mhasawade2021machine,
  title={Machine learning and algorithmic fairness in public and population health},
  author={Mhasawade, Vishwali and Zhao, Yuan and Chunara, Rumi},
  journal={Nature Machine Intelligence},
  volume={3},
  number={8},
  pages={659--666},
  year={2021},
  publisher={Nature Publishing Group UK London}
}

@article{anselin2009spatial,
  title={Spatial regression},
  author={Anselin, Luc},
  journal={The SAGE handbook of spatial analysis},
  volume={1},
  pages={255--276},
  year={2009},
  publisher={Sage Publications Los Angeles, California, USA}
}

@article{fotheringham2009geographically,
  title={Geographically weighted regression},
  author={Fotheringham, A Stewart and Brunsdon, Chris and Charlton, ME},
  journal={The Sage handbook of spatial analysis},
  volume={1},
  pages={243--254},
  year={2009},
  publisher={Sage Thousand Oaks, CA}
}

@article{ke2017lightgbm,
  title={Lightgbm: A highly efficient gradient boosting decision tree},
  author={Ke, Guolin and Meng, Qi and Finley, Thomas and Wang, Taifeng and Chen, Wei and Ma, Weidong and Ye, Qiwei and Liu, Tie-Yan},
  journal={Advances in neural information processing systems},
  volume={30},
  year={2017}
}

@article{krenz2023linking,
  title={Linking the urban environment and health: an innovative methodology for measuring individual-level environmental exposures},
  author={Krenz, Kimon and Dhanani, Ashley and McEachan, Rosemary RC and Sohal, Kuldeep and Wright, John and Vaughan, Laura},
  journal={International Journal of Environmental Research and Public Health},
  volume={20},
  number={3},
  pages={1953},
  year={2023},
  publisher={MDPI}
}

@inproceedings{klemmer2023positional,
  title={Positional encoder graph neural networks for geographic data},
  author={Klemmer, Konstantin and Safir, Nathan S and Neill, Daniel B},
  booktitle={International Conference on Artificial Intelligence and Statistics},
  pages={1379--1389},
  year={2023},
  organization={PMLR}
}

@inproceedings{kong2023leveraging,
  title={Leveraging responsible, explainable, and local artificial intelligence solutions for clinical public health in the global south},
  author={Kong, Jude Dzevela and Akpudo, Ugochukwu Ejike and Effoduh, Jake Okechukwu and Bragazzi, Nicola Luigi},
  booktitle={Healthcare},
  volume={11},
  number={4},
  pages={457},
  year={2023},
  organization={MDPI}
}

@article{molnar2020pitfalls,
  title={Pitfalls to avoid when interpreting machine learning models},
  author={Molnar, Christoph and K{\"o}nig, Gunnar and Herbinger, Julia and Freiesleben, Timo and Dandl, Susanne and Scholbeck, Christian A and Casalicchio, Giuseppe and Grosse-Wentrup, Moritz and Bischl, Bernd},
  year={2020}
}

@article{corburn2004confronting,
  title={Confronting the challenges in reconnecting urban planning and public health},
  author={Corburn, Jason},
  journal={American journal of public health},
  volume={94},
  number={4},
  pages={541--546},
  year={2004},
  publisher={American Public Health Association}
}

@article{gao2024uncertainty,
  title={Uncertainty-aware probabilistic graph neural networks for road-level traffic crash prediction},
  author={Gao, Xiaowei and Jiang, Xinke and Haworth, James and Zhuang, Dingyi and Wang, Shenhao and Chen, Huanfa and Law, Stephen},
  journal={Accident Analysis \& Prevention},
  volume={208},
  pages={107801},
  year={2024},
  publisher={Elsevier}
}

@article{raco2014urban,
  title={Urban policies on diversity in London, United Kingdom},
  author={Raco, Mike and Kesten, Jamie and Colomb, Claire},
  journal={Bartlett School of Planning, University College London, London},
  year={2014}
}

@misc{batty2018artificial,
  title={Artificial intelligence and smart cities},
  author={Batty, Michael},
  journal={Environment and Planning B: Urban Analytics and City Science},
  volume={45},
  number={1},
  pages={3--6},
  year={2018},
  publisher={SAGE Publications Sage UK: London, England}
}

@article{gilardi2021social,
  title={Social media and policy responses to the COVID-19 pandemic in Switzerland},
  author={Gilardi, Fabrizio and Gessler, Theresa and Kubli, Ma{\"e}l and M{\"u}ller, Stefan},
  journal={Swiss Political Science Review},
  volume={27},
  number={2},
  pages={243--256},
  year={2021},
  publisher={Wiley Online Library}
}

@article{gao2022regional,
  title={Regional inequalities and influencing factors of residents’ health in China: analysis from the perspective of opening-up},
  author={Gao, Guozhen and Hu, Jinmiao and Wang, Yuanyuan and Wang, Guofeng},
  journal={International Journal of Environmental Research and Public Health},
  volume={19},
  number={19},
  pages={12069},
  year={2022},
  publisher={MDPI}
}

@article{yue2024substantially,
  title={Substantially reducing global PM2. 5-related deaths under SDG3. 9 requires better air pollution control and healthcare},
  author={Yue, Huanbi and He, Chunyang and Huang, Qingxu and Zhang, Da and Shi, Peijun and Moallemi, Enayat A and Xu, Fangjin and Yang, Yang and Qi, Xin and Ma, Qun and others},
  journal={Nature Communications},
  volume={15},
  number={1},
  pages={2729},
  year={2024},
  publisher={Nature Publishing Group UK London}
}

@article{hunter2023advancing,
  title={Advancing urban green and blue space contributions to public health},
  author={Hunter, Ruth Fiona and Nieuwenhuijsen, Mark and Fabian, Carlo and Murphy, Niamh and O'Hara, Kelly and Rappe, Erja and Sallis, James Fleming and Lambert, Estelle Victoria and Duenas, Olga Lucia Sarmiento and Sugiyama, Takemi and others},
  journal={The lancet public health},
  volume={8},
  number={9},
  pages={e735--e742},
  year={2023},
  publisher={Elsevier}
}

@article{comber2011spatial,
  title={A spatial analysis of variations in health access: linking geography, socio-economic status and access perceptions},
  author={Comber, Alexis J and Brunsdon, Chris and Radburn, Robert},
  journal={International journal of health geographics},
  volume={10},
  number={1},
  pages={44},
  year={2011},
  publisher={Springer}
}

@article{shrestha2016environmental,
  title={Environmental health related socio-spatial inequalities: identifying “hotspots” of environmental burdens and social vulnerability},
  author={Shrestha, Rehana and Flacke, Johannes and Martinez, Javier and Van Maarseveen, Martin},
  journal={International journal of environmental research and public health},
  volume={13},
  number={7},
  pages={691},
  year={2016},
  publisher={MDPI}
}

@article{cheng2014spatiotemporal,
  title={Spatiotemporal data mining},
  author={Cheng, Tao and Haworth, James and Anbaroglu, Berk and Tanaksaranond, Garavig and Wang, Jiaqiu},
  journal={Handbook of regional science},
  pages={1173--1193},
  year={2014},
  publisher={Springer}
}

@article{zhao2025characterization,
  title={Characterization and estimation of heterogeneous spatial autocorrelation in spatial autoregressive models},
  author={Zhao, Jing and Pu, Yue},
  journal={PloS one},
  volume={20},
  number={7},
  pages={e0327316},
  year={2025},
  publisher={Public Library of Science San Francisco, CA USA}
}

@article{rudin2019stop,
  title={Stop explaining black box machine learning models for high stakes decisions and use interpretable models instead},
  author={Rudin, Cynthia},
  journal={Nature machine intelligence},
  volume={1},
  number={5},
  pages={206--215},
  year={2019},
  publisher={Nature Publishing Group UK London}
}

@inproceedings{caruana2015intelligible,
  title={Intelligible models for healthcare: Predicting pneumonia risk and hospital 30-day readmission},
  author={Caruana, Rich and Lou, Yin and Gehrke, Johannes and Koch, Paul and Sturm, Marc and Elhadad, Noemie},
  booktitle={Proceedings of the 21th ACM SIGKDD international conference on knowledge discovery and data mining},
  pages={1721--1730},
  year={2015}
}

@article{wirtz2021artificial,
  title={Artificial intelligence in the public sector-a research agenda},
  author={Wirtz, Bernd W and Langer, Paul F and Fenner, Carolina},
  journal={International Journal of Public Administration},
  volume={44},
  number={13},
  pages={1103--1128},
  year={2021},
  publisher={Taylor \& Francis}
}

@article{lantz1998socioeconomic,
  title={Socioeconomic factors, health behaviors, and mortality: results from a nationally representative prospective study of US adults},
  author={Lantz, Paula M and House, James S and Lepkowski, James M and Williams, David R and Mero, Richard P and Chen, Jieming},
  journal={Jama},
  volume={279},
  number={21},
  pages={1703--1708},
  year={1998},
  publisher={American Medical Association}
}

@article{apter1999influence,
  title={The influence of demographic and socioeconomic factors on health-related quality of life in asthma},
  author={Apter, Andrea J and Reisine, Susan T and Affleck, Glenn and Barrows, Erik and ZuWallack, Richard L},
  journal={Journal of Allergy and Clinical Immunology},
  volume={103},
  number={1},
  pages={72--78},
  year={1999},
  publisher={Elsevier}
}

@article{chen2008air,
  title={Air pollution and population health: a global challenge},
  author={Chen, Bingheng and Kan, Haidong},
  journal={Environmental health and preventive medicine},
  volume={13},
  number={2},
  pages={94--101},
  year={2008},
  publisher={Springer}
}

@article{kjellstrom2007urban,
  title={Urban environmental health hazards and health equity},
  author={Kjellstrom, Tord and Friel, Sharon and Dixon, Jane and Corvalan, Carlos and Rehfuess, Eva and Campbell-Lendrum, Diarmid and Gore, Fiona and Bartram, Jamie},
  journal={Journal of urban health},
  volume={84},
  number={Suppl 1},
  pages={86--97},
  year={2007},
  publisher={Springer}
}

@article{nutsford2013ecological,
  title={An ecological study investigating the association between access to urban green space and mental health},
  author={Nutsford, Daniel and Pearson, Amber L and Kingham, Simon},
  journal={Public health},
  volume={127},
  number={11},
  pages={1005--1011},
  year={2013},
  publisher={Elsevier}
}

@article{marmot2005social,
  title={Social determinants of health inequalities},
  author={Marmot, Michael},
  journal={The lancet},
  volume={365},
  number={9464},
  pages={1099--1104},
  year={2005},
  publisher={Elsevier}
}

@article{fecht2015associations,
  title={Associations between air pollution and socioeconomic characteristics, ethnicity and age profile of neighbourhoods in England and the Netherlands},
  author={Fecht, Daniela and Fischer, Paul and Fortunato, L{\'e}a and Hoek, Gerard and De Hoogh, Kees and Marra, Marten and Kruize, Hanneke and Vienneau, Danielle and Beelen, Rob and Hansell, Anna},
  journal={Environmental pollution},
  volume={198},
  pages={201--210},
  year={2015},
  publisher={Elsevier}
}

@article{bixby2015associations,
  title={Associations between green space and health in English cities: an ecological, cross-sectional study},
  author={Bixby, Honor and Hodgson, Susan and Fortunato, L{\'e}a and Hansell, Anna and Fecht, Daniela},
  journal={PLoS One},
  volume={10},
  number={3},
  pages={e0119495},
  year={2015},
  publisher={Public Library of Science San Francisco, CA USA}
}

@article{twohig2018health,
  title={The health benefits of the great outdoors: A systematic review and meta-analysis of greenspace exposure and health outcomes},
  author={Twohig-Bennett, Caoimhe and Jones, Andy},
  journal={Environmental research},
  volume={166},
  pages={628--637},
  year={2018},
  publisher={Elsevier}
}

@article{velivckovic2018deep,
  title={Deep graph infomax},
  author={Velivckovi{\'c}, Petar and Fedus, William and Hamilton, William L and Li{\`o}, Pietro and Bengio, Yoshua and Hjelm, R Devon},
  journal={arXiv preprint arXiv:1809.10341},
  year={2018}
}

@article{brody2021attentive,
  title={How attentive are graph attention networks?},
  author={Brody, Shaked and Alon, Uri and Yahav, Eran},
  journal={arXiv preprint arXiv:2105.14491},
  year={2021}
}

@article{wang2025multi,
  title={Multi-modal contrastive learning of urban space representations from POI data},
  author={Wang, Xinglei and Cheng, Tao and Law, Stephen and Zeng, Zichao and Yin, Lu and Liu, Junyuan},
  journal={Computers, Environment and Urban Systems},
  volume={120},
  pages={102299},
  year={2025},
  publisher={Elsevier}
}

@article{grekousis2025geographical,
  title={Geographical-XGBoost: a new ensemble model for spatially local regression based on gradient-boosted trees},
  author={Grekousis, George},
  journal={Journal of Geographical Systems},
  pages={1--27},
  year={2025},
  publisher={Springer}
}

@article{li2022extracting,
  title={Extracting spatial effects from machine learning model using local interpretation method: An example of SHAP and XGBoost},
  author={Li, Ziqi},
  journal={Computers, Environment and Urban Systems},
  volume={96},
  pages={101845},
  year={2022},
  publisher={Elsevier}
}

@article{nikparvar2021machine,
  title={Machine learning of spatial data},
  author={Nikparvar, Behnam and Thill, Jean-Claude},
  journal={ISPRS International Journal of Geo-Information},
  volume={10},
  number={9},
  pages={600},
  year={2021},
  publisher={MDPI}
}

@inproceedings{shao2015community,
  title={Community detection based on distance dynamics},
  author={Shao, Junming and Han, Zhichao and Yang, Qinli and Zhou, Tao},
  booktitle={Proceedings of the 21th ACM SIGKDD international conference on knowledge discovery and data mining},
  pages={1075--1084},
  year={2015}
}

@misc{malleson2024digital,
  title={Digital twins on trial: Can they actually solve wicked societal problems and change the world for better?},
  author={Malleson, Nick and Franklin, Rachel and Arribas-Bel, Daniel and Cheng, Tao and Birkin, Mark},
  journal={Environment and Planning B: Urban Analytics and City Science},
  volume={51},
  number={6},
  pages={1181--1186},
  year={2024},
  publisher={SAGE Publications Sage UK: London, England}
}

@inproceedings{birks2020towards,
  title={Towards the development of societal twins},
  author={Birks, Dan and Heppenstall, Alison and Malleson, Nick},
  booktitle={Frontiers in Artificial Intelligence and Applications},
  volume={325},
  pages={2883--2884},
  year={2020},
  organization={Leeds}
}

@article{xu2018powerful,
  title={How powerful are graph neural networks?},
  author={Xu, Keyulu and Hu, Weihua and Leskovec, Jure and Jegelka, Stefanie},
  journal={arXiv preprint arXiv:1810.00826},
  year={2018}
}

@article{brown2025alphaearth,
  title={Alphaearth foundations: An embedding field model for accurate and efficient global mapping from sparse label data},
  author={Brown, Christopher F and Kazmierski, Michal R and Pasquarella, Valerie J and Rucklidge, William J and Samsikova, Masha and Zhang, Chenhui and Shelhamer, Evan and Lahera, Estefania and Wiles, Olivia and Ilyushchenko, Simon and others},
  journal={arXiv preprint arXiv:2507.22291},
  year={2025}
}

\clearpage
\appendix
\section{Feature Selection Methodology}\label{app:feature_selection}

This appendix provides detailed information about the feature selection process used to reduce the dimensionality of the MedSAT dataset from 155 variables to 67 variables.

\begin{figure}[H]
    \centering
    \includegraphics[width=\textwidth]{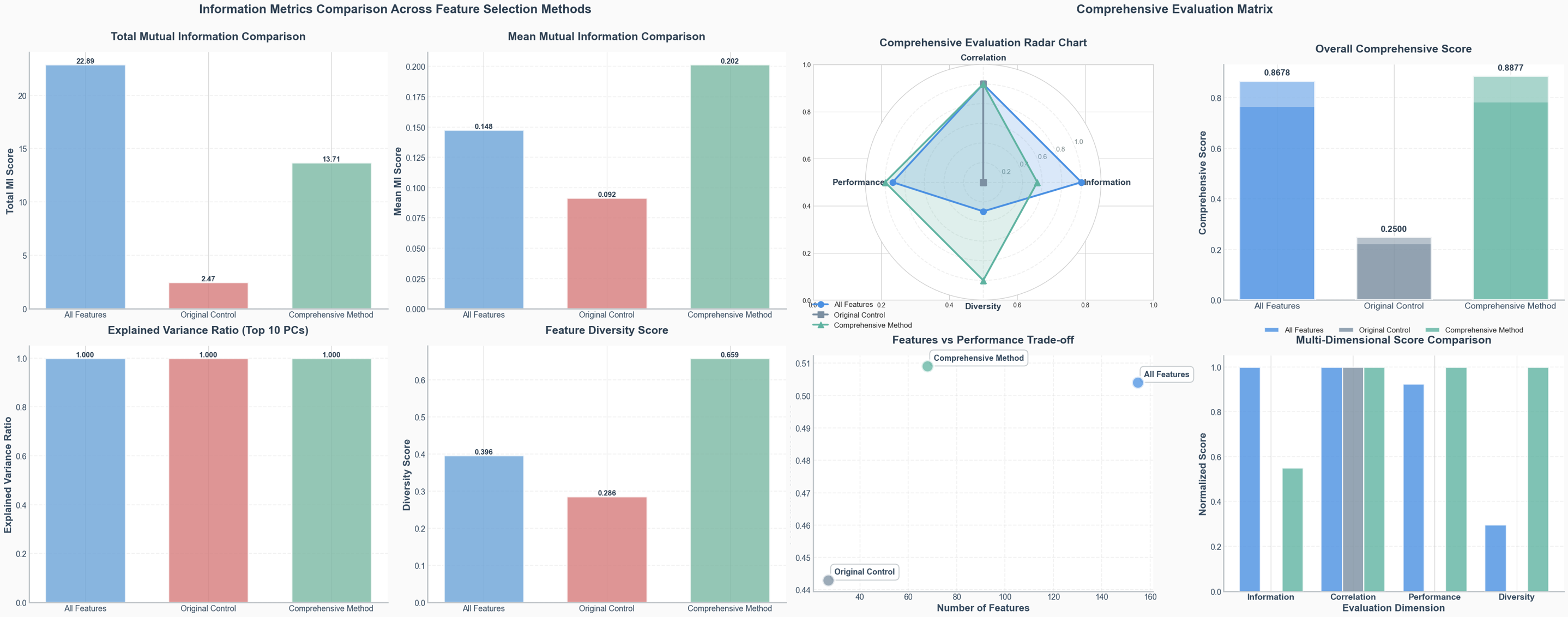}
    \caption{Comprehensive evaluation of feature selection strategies. Panels summarise: (top-left) total mutual information (MI) of retained features; (top-middle) mean MI; (top-right) radar plot comparing four normalised dimensions (information, correlation control, performance, diversity); (top-far-right) overall composite score. Bottom row: (left) explained variance ratio (EVR) of the top 10 principal components; (middle-left) diversity score measuring category coverage; (middle-right) trade-off between number of features and performance; (right) per-dimension normalised scores. The ``Comprehensive Method'' balances information content and diversity while maintaining high performance with a compact feature set, improving over the naive ``All Features'' and the minimal ``Original Control'' set.}
    \label{fig:feature_selection_matrix}
\end{figure}

Figure~\ref{fig:feature_selection_matrix} compares three strategies evaluated in this study: using all variables (\emph{All Features}), a minimal confounder set (\emph{Original Control}), and our multi-criteria pipeline (\emph{Comprehensive Method}). The latter maximises information (total/mean MI) under explicit correlation control and achieves near-saturated EVR with markedly fewer variables, yielding higher composite scores and a more diverse, policy-relevant feature portfolio. This selection underpins the stability of UST-GNN and the clarity of subsequent interpretability analyses.

\subsection{Comprehensive Multi-Dimensional Feature Selection Strategy}

Our feature selection strategy prioritized theoretical relevance and causal inference robustness while leveraging data-driven insights to optimize predictive performance. We selected 67 variables from 155 available features through a systematic multi-dimensional evaluation framework that integrates domain knowledge with statistical learning approaches, ensuring comprehensive coverage of socio-demographic, environmental, and behavioral determinants of health outcomes.

The selection process comprised four key stages:

\textbf{(i) Multi-dimensional feature scoring:} We computed feature importance scores using four complementary evaluation metrics: mutual information (MI) regression, F-statistics from univariate linear regression, random forest feature importances, and Lasso regularization coefficients. Each metric captured different aspects of feature–outcome relationships: MI quantified non-linear associations, F-statistics assessed linear dependencies, random forest importances captured interactive effects, and Lasso coefficients reflected sparsity-regularized contributions. Individual scores were normalized to $[0,1]$ and combined using weighted averaging (weights: 0.3 for MI, 0.2 for F-statistics, 0.3 for random forest, and 0.2 for Lasso), producing a unified \textit{combined score} that balanced multiple statistical perspectives.

\textbf{(ii) Adaptive feature number determination:} Rather than fixing a predefined feature count, we applied cross-validation (5-fold) over candidate feature set sizes ($k \in [30, 80]$) to identify the optimal dimensionality that maximized predictive performance while maintaining model stability. For each candidate $k$, we selected the top-$k$ features ranked by combined scores and evaluated their collective performance using Random Forest regression with cross-validation $R^2$ as the selection criterion.

\textbf{(iii) Mandatory control variable inclusion:} Core control variables commonly used as key confounders in public health and broader socio-spatial research, including age distributions, gender, ethnicity, occupation types, population density, and key environmental exposures (NDVI, NO\textsubscript{2}, land cover types)—were automatically retained regardless of their combined scores. This ensured that theoretically essential predictors remained represented across all model configurations. Because a small set of theory-driven control variables was mandatorily retained, some residual collinearity may remain in the final feature set. This reflects a deliberate trade-off between strict redundancy removal and the preservation of substantively important controls.

\textbf{(iv) Redundancy elimination:} To address multicollinearity while preserving feature diversity, we computed pairwise correlation matrices for selected features and identified highly correlated pairs ($|r| > 0.8$). For each pair, we retained the feature with the higher combined score (or prioritized control variables when one feature was a mandatory control), removing redundant predictors without substantial information loss.
\begin{figure}[H]
    \centering
    \includegraphics[width=\textwidth]{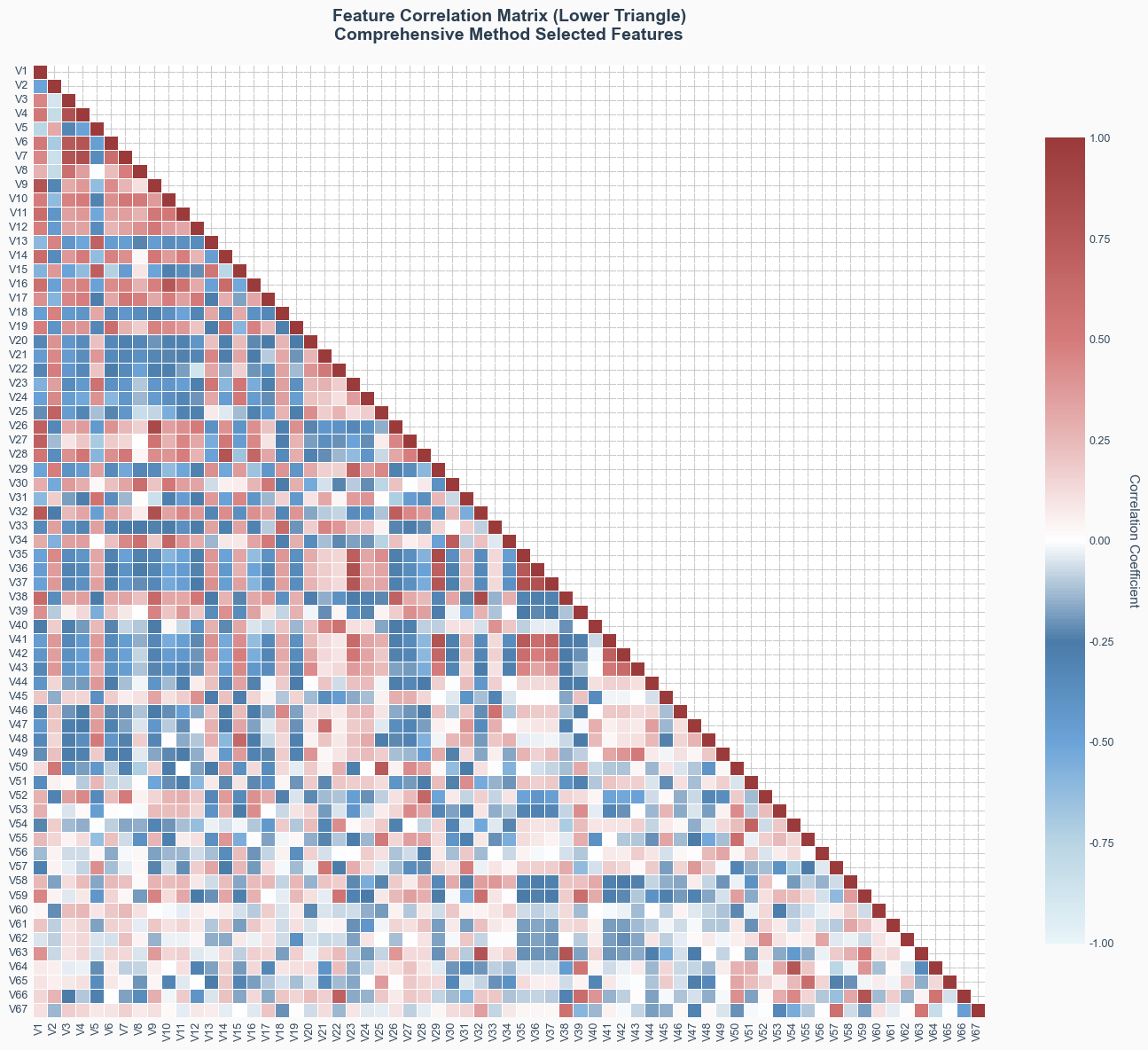}
    \caption{
    Pairwise correlation matrix (lower triangle) of the 67 variables retained by the comprehensive feature selection method. 
    Each cell represents the Pearson correlation coefficient between two features (red: positive; blue: negative). 
    The pattern illustrates that strong pairwise dependencies ($|r|>0.8$) were effectively minimized through the redundancy elimination step, resulting in a feature set that maintains informational diversity across socio-demographic, environmental, and behavioural dimensions.}
    \label{fig:feature_corr_matrix}
\end{figure}

This multi-stage approach balances dimensionality reduction with the retention of sufficient feature diversity to capture complex dependencies, thereby supporting both model stability and interpretable embedding–feature associations. The final feature set comprised 67 variables (54 socio-demographic and 13 environmental), representing an optimal balance between comprehensiveness and parsimony.
Figure~\ref{fig:feature_corr_matrix} visualizes the pairwise correlation structure among the 67 variables selected by our comprehensive multi-dimensional feature selection framework. The lower-triangular correlation map demonstrates that the retained variables exhibit a balanced mix of moderate associations without strong linear redundancy. Clusters of moderate correlations (notably within age-related or occupation-related variables) indicate internally coherent thematic groups, yet the overall absence of dense high-correlation blocks ($|r|>0.8$).

\subsection{Selected Variables Summary}

Table~\ref{tab:selected_variables_appendix} presents the complete list of 67 selected variables, organized by category. The selection encompasses key demographic characteristics (age distribution, gender, ethnicity), socioeconomic indicators (occupation, housing, deprivation), and environmental exposures (air quality, greenness, land cover).

\begin{table}[h]
\caption{Complete list of 67 selected variables organized by category.}
\label{tab:selected_variables_appendix}
\centering
\footnotesize
\setlength{\tabcolsep}{2pt}
\renewcommand{\arraystretch}{0.85}
\begin{tabular}{p{6cm} p{8cm}}
\toprule
\textbf{Category} & \textbf{Selected Variables} \\
\midrule
\textbf{Demographics} & Age groups (10-14, 15-19, 20-24, 25-29, 30-34, 35-39, 40-44, 45-49, 50-54, 55-59, 60-64, 65-69, 70-74, 75-79, 80-84, 85+), Male proportion \\
\textbf{Ethnicity} & Mixed race, White, Asian, Black populations \\
\textbf{Socioeconomic} & Professional occupations, Population density, Household deprivation \\
\textbf{Environment} & NO\textsubscript{2}, NDVI, Water bodies, Trees, Grass, Bare land \\
\textbf{Housing} & Central heating availability, Communal heating, Occupancy ratings \\
\textbf{Transport} & Commute modes (foot, metro rail, bus, bicycle, train), Distance to work \\
\textbf{Employment} & Working hours, Part-time work, Work from home, Student population \\
\textbf{Residential} & Length of residence, Marital status, English proficiency \\
\textbf{Religion} & Christian, Buddhist, Muslim populations \\
\textbf{Climate} & Lake depth, Snow cover, Surface runoff, Snow/ice, Crops, Canopy evaporation, Shrub cover \\
\bottomrule
\end{tabular}
\end{table}

\section{Interpretability}\label{app:explainability}
\begin{figure}[!h]
    \centering
    \includegraphics[width=0.98\textwidth]{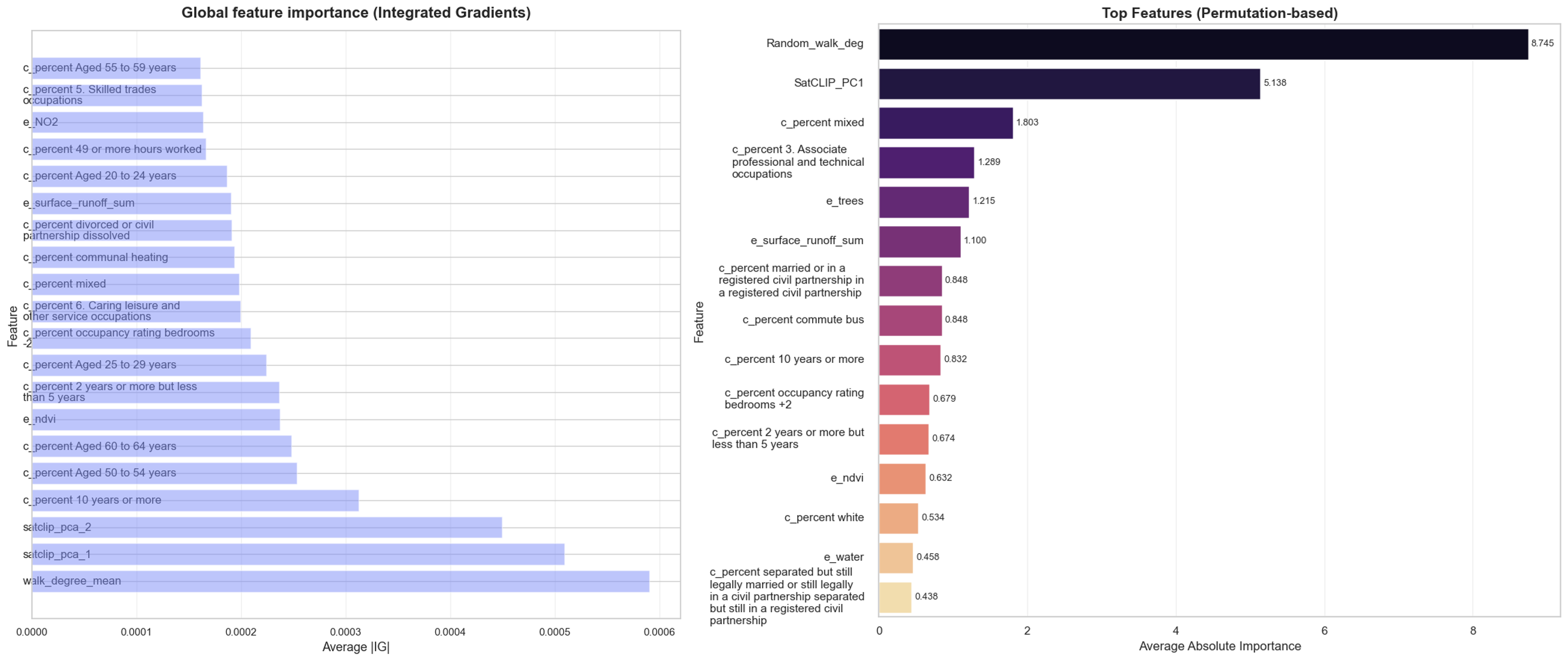}
    \caption{Permutation-based global feature importance. Bars show performance degradation after shuffling each feature, summarising city-wide contribution. Labels are wrapped for readability.}
    \label{fig:perm_importance}
\end{figure}

\begin{figure}[!h]
    \centering
    \includegraphics[width=0.98\textwidth]{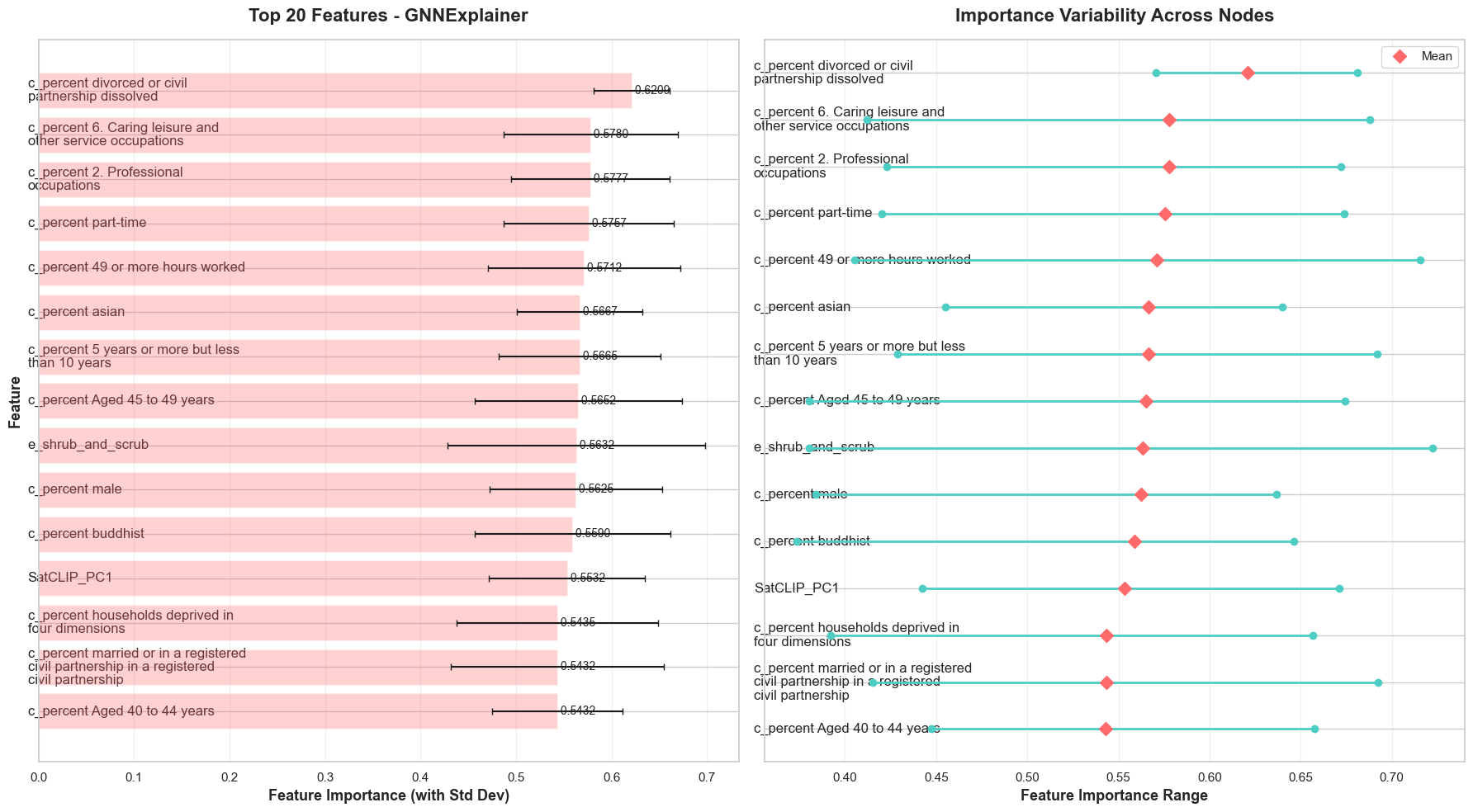}
    \caption{GNNExplainer-based summaries over five representative nodes (10th--90th percentiles). Left: bar chart with uncertainty; Right: range of importance across nodes. Stochastic optimisation and sparse sampling can yield local deviations from global rankings.}
    \label{fig:gnnexplainer_based}
\end{figure}
\subsection{Motivation and Evaluation Criteria}
Interpreting spatial graph models requires balancing methodological faithfulness and practical communicability. In the context of urban analytics, meaningful explanations should (i) remain faithful to the trained model, (ii) exhibit stability under random seeds and data perturbations, (iii) incorporate spatial topology rather than treating nodes independently, and (iv) provide outputs interpretable to domain experts such as planners or epidemiologists. Guided by these principles, we evaluate and compare four complementary interpretability families—embedding PCA (ours), gradient-based saliency, permutation importance, and GNNExplainer—each representing a different trade-off between global generality and local fidelity.

\subsection{PCA-Embedding (ours)}
Our PCA-embedding interpretation departs from traditional attribution frameworks by focusing on the structure of the learned representation rather than direct input–output gradients. Applying PCA to the final-layer node embeddings enables us to identify dominant axes of variation that capture the macro-scale organisation of urban socio-environmental patterns. The approach is architecture-agnostic, computationally efficient, and stable across folds, producing interpretable spatial gradients that link \textit{raw covariates, learned embeddings, and observed outcomes} within a single analytic space. 

Although PCA does not assign additive feature importances and may blend correlated influences, its inductive and correlation-oriented nature makes it particularly suited to complex urban systems where strict causal decomposition is rarely identifiable. By summarising high-dimensional latent spaces into continuous spatial surfaces, the PCA method transforms model internal representations into city-scale narratives—revealing coherent gradients and clusters that conventional attribution methods fail to convey.

\subsection{Gradient-Based and Integrated Gradients Methods}
Gradient-based saliency and Integrated Gradients (IG) provide node-level sensitivity analyses that are directly faithful to model derivatives. They excel at highlighting how small input perturbations influence predictions and are theoretically grounded (IG satisfies axioms such as completeness). However, in graph contexts these gradients entangle self-node and neighbour effects, complicating their aggregation into global or spatially coherent interpretations. In our experiments, IG maps yielded consistent but spatially fragmented patterns, confirming their strength in local attribution but limited communicability at city scale.
\subsection{Permutation-Based Importance}
Permutation-based importance evaluates how shuffling each feature affects model performance, offering a model-agnostic and intuitively “what-if” assessment of feature relevance. The method produced stable global rankings aligned with known socio-environmental correlates (e.g., population density, greenery, occupation type), corroborating PCA-derived associations. Yet, because it ignores graph topology and can inflate correlated variables’ importance, permutation analysis complements but cannot substitute embedding-level structure discovery.
\subsection{GNNExplainer}
GNNExplainer directly optimises a sparse subgraph and feature mask that most strongly influence a node’s prediction. This yields rich local explanations incorporating topology, suitable for case-level insights or motif discovery. Nevertheless, its stochastic optimisation introduces variability across runs, and aggregating multiple local explanations into a consistent global interpretation is computationally intensive. As shown in Figure~\ref{fig:gnnexplainer_based}, feature rankings fluctuate across nodes, illustrating the challenge of scaling local explainers to city-wide contexts.
\subsection{Cross-Method Comparison and Discussion}
Figure~\ref{fig:perm_importance} and Figure~\ref{fig:gnnexplainer_based} summarise the three feature-level explanation methods, while Figure~\ref{fig:pca_map} (main text) presents the embedding-level PCA interpretation. Across methods, key socio-demographic and environmental variables—such as age structure, occupation type, commuting mode, greenness, and ethnic composition—consistently emerge as explanatory factors, affirming the substantive coherence of the learned spatial patterns.  

In essence, PCA-embedding interpretation reframes interpretability from an attribution problem to a structural one—seeking to understand how the model organises information across space rather than merely how it responds to feature perturbations. This paradigm aligns more naturally with the goals of urban science, where understanding spatial heterogeneity and systemic dependencies often outweighs pinpointing individual variable causality.

\paragraph{Summary}
PCA offers a transparent, stable, and computationally light interpretability pathway, revealing global structure–function relationships that complement the local precision of gradient and GNNExplainer methods. Together, these complementary techniques form a triangulated interpretability framework: PCA provides macro-scale coherence, gradients ensure model faithfulness, permutation tests assess robustness, and GNNExplainer yields local topology awareness. Their convergence substantiates that UST-GNN captures genuine, spatially embedded mechanisms rather than artefacts of data correlation or model leakage.

\paragraph{Takeaway} Our PCA-embedding method excels at revealing fold-stable, spatially coherent gradients and relating them to socio–environmental variables, offering a compact, communicable narrative. Gradient/IG and permutation importance provide faithful, feature-level corroboration; GNNExplainer offers case-level, topology-aware insight.

\section{Robustness and Reliability Tests}\label{app:robustness}
\subsection{Fast permutation test (OOF-based)}
We conducted a fast permutation test using out-of-fold (OOF) predictions to assess whether the observed performance could arise by chance. Holding the OOF predictions fixed and randomly permuting the ground-truth labels 1{,}000 times produced a null distribution of $R^2$ centred at $-0.8819$ with standard deviation $0.0093$. The observed OOF $R^2$ was $0.926$, yielding a permutation $p$-value of $<0.001$ (resolution-limited by 1{,}000 permutations). The complete separation between the observed value and the null confirms that the model captures genuine structure rather than accidental fit.

\subsection{Lightweight learning curve}
To verify generalisation with reduced runtime, we computed a lightweight learning curve using five training sizes (20--100\%), capped epochs, and a fixed test fold. Test $R^2$ increased monotonically with training size (approx.
0.39 $\to$ 0.83), while train $R^2$ decreased towards the test curve, indicating reduced variance and healthy capacity control. Together with the permutation test, these results support the reliability and robustness of UST-GNN.

% \section{Open-Source Repository and Supporting Materials}\label{app:open_source_release}

\end{document}